\documentclass[remotesensing,article,accept,pdftex,moreauthors]{Definitions/mdpi}

\firstpage{1} 
\pubvolume{17}
\issuenum{16}
\articlenumber{2845}
\pubyear{2025}
\copyrightyear{2025}
\externaleditor{Li Fang, Jian Yang, Yu Feng} 
\datereceived{21 May 2025} 
\daterevised{20 July 2025} 
\dateaccepted{29 July 2025} 
\datepublished{15 August 2025} 
\hreflink{https://doi.org/10.3390/rs17162845} 

\usepackage{pifont}%
\usepackage{xcolor}
\newcommand{\cmark}{\ding{51}}%
\newcommand{\xmark}{\ding{55}}%

\Title {Lane Graph Extraction from Aerial Imagery via Lane Segmentation Refinement with Diffusion Models}

\TitleCitation{Lane Graph Extraction from Aerial Imagery via Lane Segmentation Refinement with Diffusion Models}

\Author{Antonio Ruiz 
 $^{1,2,}$\orcidA{}, Andrew Melnik $^{1,2,\dagger}$, Nicolo Savioli $^{1}$, Dong Wang $^{1}$, Yanfeng Zhang $^{1, 
 *}$ and Helge Ritter $^{2}$}

\isAPAStyle{%
       \AuthorCitation{Lastname, F., Lastname, F., \& Ritter, H.}
         }{%
        \isChicagoStyle{%
        \AuthorCitation{Lastname, Firstname, Firstname Lastname, and Firstname Lastname.}
        }{
        \AuthorCitation{Ruiz, A.; Melnik, A.; Savioli, N.; Wang, D.; Zhang, Y.; Ritter, H.}
        }
}

\address{%
$^{1}$ \quad Riemann {Lab}, Huawei, 80992 Munich, Germany; aruiz@techfak.uni-bielefeld.de (A.R.) 
\\
$^{2}$ \quad Center for Cognitive Interaction Technology (CITEC), Faculty of Technology, Bielefeld University, 33619 Bielefeld, Germany 
}

\corres{Correspondence: zhangyanfeng8@huawei.com
}

\firstnote{This work was done while Andrew was working at Bielefeld University and at Huawei.}  

\abstract{The lane graph is critical for applications such as autonomous driving and lane-level route planning. While previous research has focused on extracting lane-level graphs from aerial imagery using convolutional neural networks (CNNs) followed by post-processing segmentation-to-graph algorithms, these methods often face challenges in producing sharp and complete segmentation masks. Challenges such as occlusions, variations in lighting, and changes in road texture can lead to incomplete and inaccurate lane masks, resulting in poor-quality lane graphs. To address these challenges, we propose a novel approach that refines the lane masks, output by a CNN, using diffusion models.
Experimental results on a publicly available dataset demonstrate that our method outperforms existing methods based solely on CNNs or diffusion models, particularly in terms of graph connectivity. Our lane mask refinement approach enhances the quality of the extracted lane graph, yielding gains of approximately 1.5\% in GEO F1 and 3.5\% in TOPO F1 scores over the best-performing CNN-based method, and improvements of 28\% and 34\%, respectively, compared to a prior diffusion-based approach. Both GEO F1 and TOPO F1 scores are critical metrics for evaluating lane graph quality. Additionally, ablation studies are conducted to evaluate the individual components of our approach, providing insights into their respective contributions and effectiveness.
}


\begin{document}


\section{Introduction}
\label{sec:introduction}

The lane graph is a critical component in autonomous driving and advanced driver-assistance systems (ADAS). It provides a structured representation of the road environment, modeling lanes as directed edges and representing intersections, merging points, and splitting points as nodes, effectively capturing the connectivity, directionality, and spatial relationships among lanes. Each node contains precise geospatial information, facilitating accurate vehicle localization, while edges support motion planning. Together, these elements enable some level of autonomous driving and detailed navigation planning.

The lane graph can be constructed and updated by collecting data with an ego-vehicle equipped with multiple sensors such as LiDAR, cameras, and inertial measurement units (IMUs). However, this approach has several limitations, including a restricted field of view, long data collection times, and dependence on vehicle-based infrastructure, which also increases costs. By contrast, aerial imagery can rapidly cover larger areas and eliminates the need of vehicle-based infrastructure, resulting in a more efficient and scalable solution compared to sensor-equipped ego-vehicles. Nevertheless, constructing an accurate lane graph from aerial imagery remains challenging due to several factors, including the large area to be covered, the complexity of crossroads, the low resolution of ground objects, changes in road texture, occlusions caused by trees, vehicle queues or bridges, and variations in lighting conditions caused by shadows.

It is also worth mentioning that the task of lane graph extraction differs from lane line detection~\cite{yao2024building} and lane marking detection~\cite{azimi2018aerial}. Lane graph extraction focuses on the topology of lane centerlines and how lanes are interconnected, whereas lane line detection targets the boundaries of the lanes, and lane marking detection focuses solely on identifying markings without considering the topological relationships between lanes.

Approaches for extracting lane graphs from aerial imagery can be broadly categorized into two main types: segmentation-based methods~\cite{he2022lane} and graph-based methods~\cite{buchner2023learning, blayney2024bezier}. Segmentation-based methods start by predicting lane segmentation masks, which are subsequently processed using conventional segmentation-to-graph algorithms to obtain the final lane graph. In contrast, graph-based methods directly predict the lane graph without relying on segmentation-to-graph algorithms. These methods may employ an agent~\cite{buchner2023learning} to predict local lane graphs based on its position or use parametric graph representations such as Bézier graphs~\cite{blayney2024bezier}. There are trade-offs between the two approaches. Segmentation-based methods offer faster and more consistent inference times as the area increases, making them more scalable. However, they are highly dependent on the quality of the lane segmentation masks to produce  high-quality lane graphs and often struggle with complex crossroads. In contrast, graph-based methods may generate more precise lane-level graphs but require algorithms to aggregate local graphs, which can significantly slow performance when dealing with large areas.

\begin{figure}[H]
  \includegraphics[width=\textwidth]{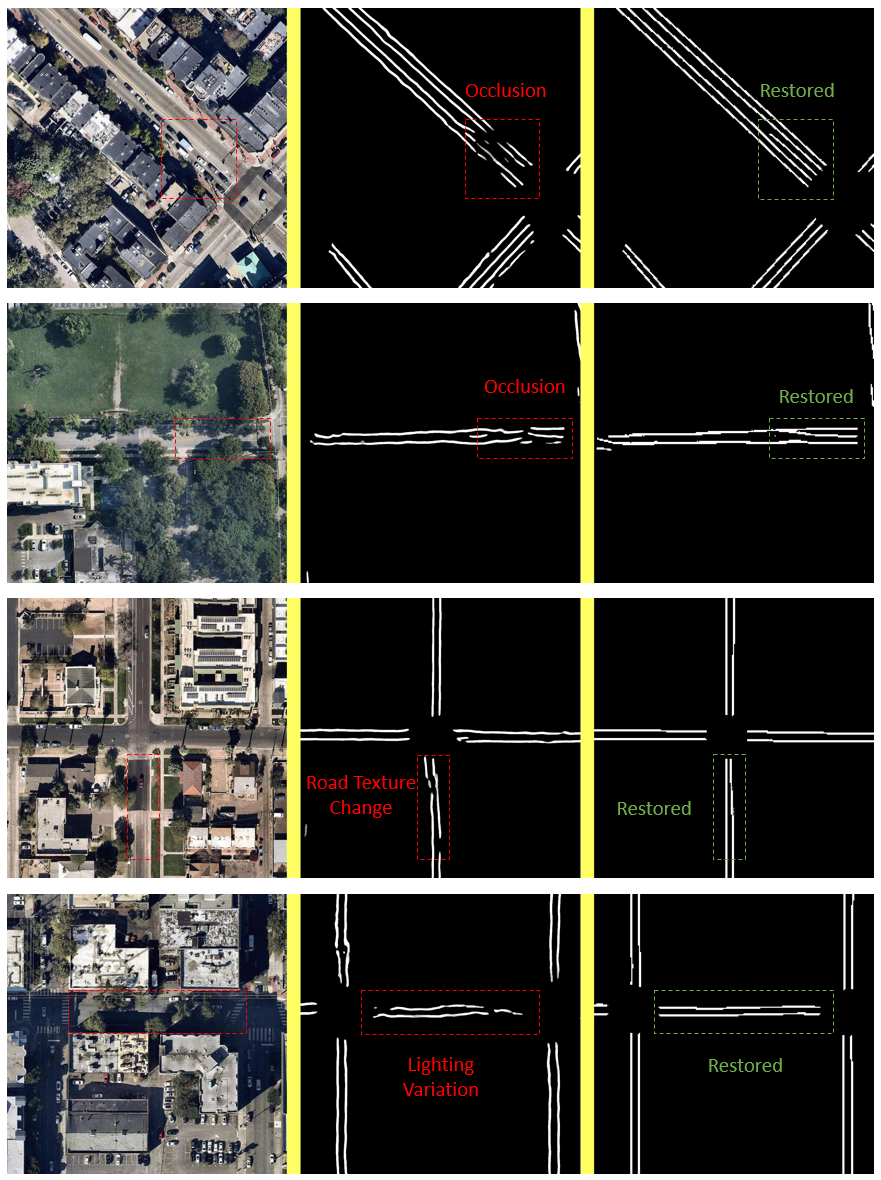}
  \caption{{Challenging 
  scenarios for CNNs. The first column shows aerial image patches containing challenging scenarios highlighted by red dotted boxes. The second column illustrates the segmentation masks produced by LaneSegmentation~\cite{he2022lane} (a CNN-based model) 
  , while the third column displays the masks predicted by our model. The first two rows highlight regions affected by occlusion, caused by queues of cars in the first row and by trees in the second row. The third row depicts a case involving a change in road texture, and the fourth illustrates the impact of lighting variation. In all these scenarios, CNNs struggle to accurately segment the lanes, whereas our model consistently produces sharp and complete lane segmentation masks.}}
  \label{fig:complex-cases}
\end{figure}

Segmentation-based methods employ CNNs to extract lane segmentation masks. However, these networks often struggle to produce sharp and continuous lane masks due to the intrinsic characteristics of aerial imagery, such as occlusions caused by trees, vehicle queues, or bridges, or changes in visual patterns resulting from variations in road texture and lighting conditions (see Figure \ref{fig:complex-cases}
). Since the subsequent segmentation-to-graph algorithm relies heavily on the quality of these masks, any inaccuracies can significantly degrade the overall quality of the lane graphs. These methods attempt to address these challenges by applying post-processing heuristics or incorporating specialized loss functions and dilated convolutional layers~\cite{he2022lane}. Nonetheless, the lane masks produced by these methods still lack continuity and sharpness. Our method addresses these issues by refining the lane segmentation masks output by a CNN, enhancing their continuity and sharpness (as shown in Figure \ref{fig:complex-cases}
), which leads to higher-quality lane graphs.

Recently, diffusion models~\cite{sohl2015deep, ho2020denoising, song2020denoising} have emerged as an alternative to solve segmentation problems, demonstrating state-of-the-art results in specific domains~\cite{amit2021segdiff, wolleb2022diffusion, wu2022medsegdiff, wu2023medsegdiff}. Diffusion models offer several advantages over CNNs, including their flexibility to create ensembles by using different seeds at inference time and ability to infer segmentation masks from complex visual patterns where CNNs often struggle. Thanks to this ability to handle complex visual patterns, we employ diffusion models to refine the lane segmentation masks. Methods~\cite{amit2021segdiff, wolleb2022diffusion, wu2022medsegdiff, wu2023medsegdiff} that leverage diffusion models for segmentation problems depend on ensembles of these models to improve the quality of segmentation mask predictions beyond what could be achieved with a single diffusion model. However, this approach falls short when segmenting thin structures, which is a critical challenge in the context of lane segmentation from aerial imagery. Slight lateral variations in the segmented pixels across the ensemble lead to a blurred average segmentation mask (refer to the discussion in Section~\ref{sec:diffusion-models-for-image-segmentation} for more details). On the other hand, a standard diffusion model, i.e., one that starts inference from Gaussian noise, lacks the robustness needed to produce high-quality segmentation masks (as confirmed and discussed in Section \ref{sec:quantitative-results}). 

\begin{figure}[H]
  \includegraphics[width=\textwidth]{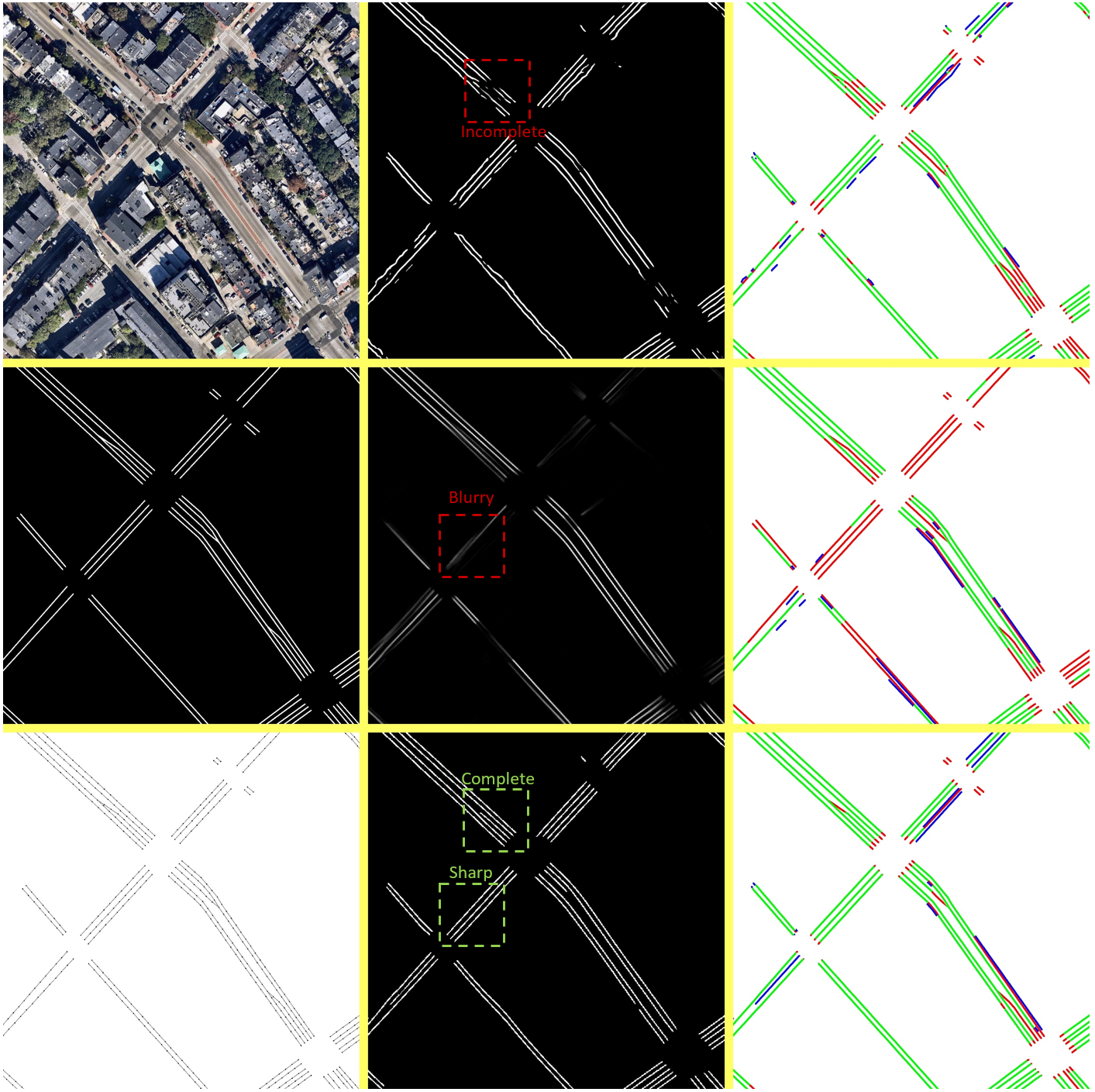}
  \caption{Visual 
  results for a region of testing tile A, comparing the outputs of the following methods: (1) LaneExtraction~\cite{he2022lane} (top row), (2) an ensemble of diffusion models~\cite{wu2022medsegdiff} (middle row), and (3) our method (bottom row). The first column shows the input aerial RGB image (top), ground truth segmentation mask (middle), and ground truth lane graph (bottom); the second and third columns display predicted lane segmentation masks and their corresponding lane graphs (used for computing the metrics). In the lane graphs (third column), green nodes indicate matched nodes, blue nodes represent false positives, and red nodes false negatives. Nodes appear as short line segments due to close spacing. Our method exhibits improved topological continuity and sharper lane segments compared to baselines.}
  \label{fig:results-tile-A}
\end{figure}

\begin{figure}[H]
  \includegraphics[width=\textwidth]{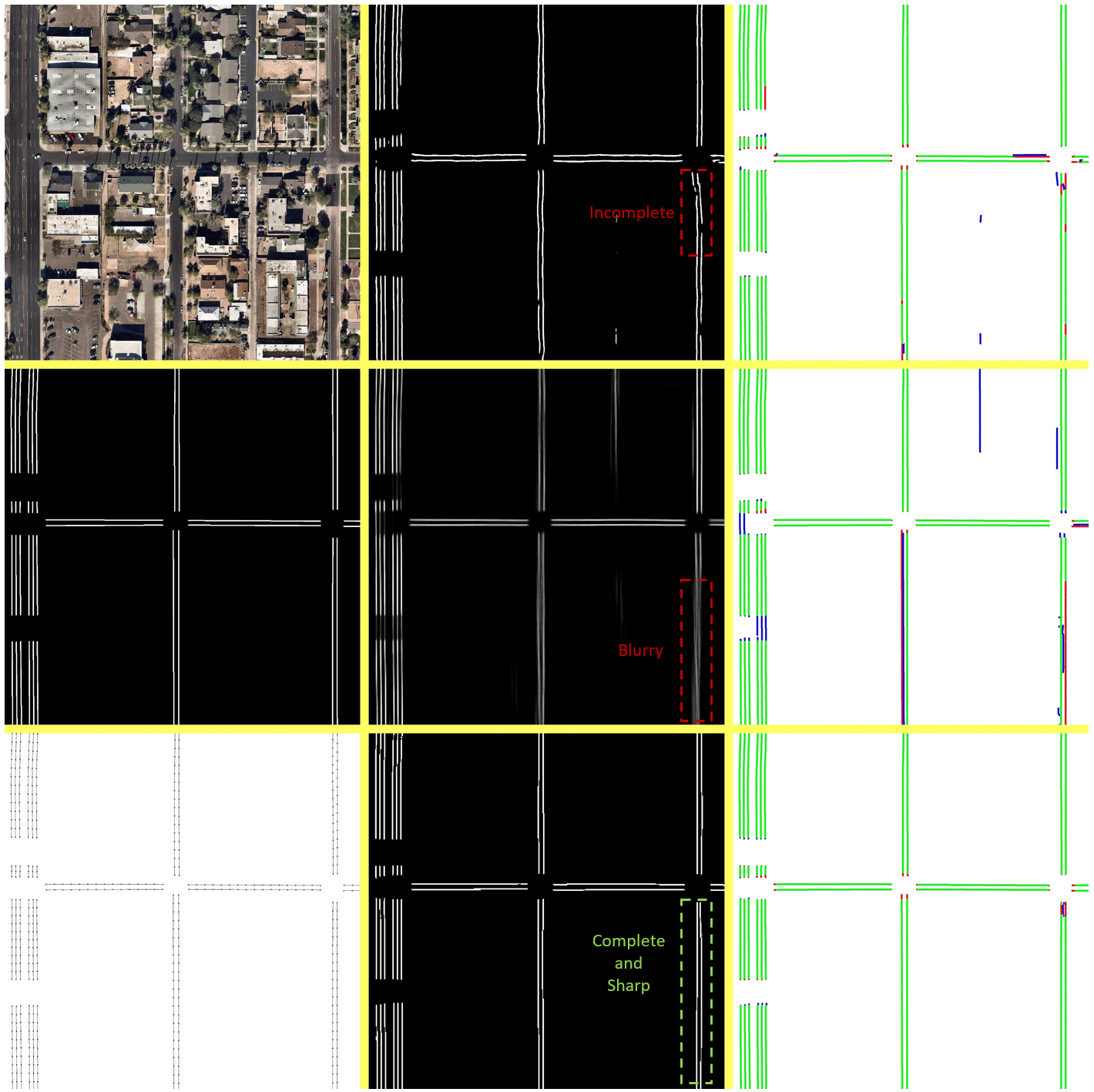}
  \caption{Visual 
  results for a region of testing tile B. Same arrangement as for tile A.}
  \label{fig:results-tile-B}
\end{figure}

To overcome these limitations, our approach avoids the ensemble strategy and employs a single diffusion model with a deterministic sampling path and a conditioned starting latent variable. Instead of initiating the sampling procedure from Gaussian noise, as it is standard in diffusion models, we start with a latent variable conditioned on the segmentation masks produced by a CNN and use a deterministic sampling. Since this starting point is already a reasonable approximation of the ground truth segmentation mask, the diffusion model improves the approximation, effectively refining the segmentation mask. A notable outcome of this refinement is that the lane masks become complete and sharp (see qualitative results in Figures 
~\ref{fig:results-tile-A} and~\ref{fig:results-tile-B}), significantly improving the connectivity of the lane graph, an essential and highly desirable attribute.

In summary, this paper makes the following contributions:

\begin{itemize}

    \item We introduce a novel approach for refining lane segmentation masks produced by a CNN. Our method employs a diffusion model with a deterministic sampling procedure, initialized from a latent variable conditioned on the initial CNN's segmentation mask predictions. These refined masks are then utilized to extract the lane graph using a conventional segmentation-to-graph algorithm. Experiments conducted on a public dataset confirm that our method, which integrates a CNN and a diffusion model, outperforms each component used individually (see Table~\ref{tab:graph-metrics}).

    \item We also carry out an ablation study on the components of the diffusion model used in our method (see Table \ref{tab:components-ablation}), as well as on the impact of varying the number of sampling steps required for lane mask refinement (see Table \ref{tab:noise-ablation}). Our results demonstrate that high-quality refinement can be achieved within only a few sampling steps. 
    
    \item Furthermore, we conduct extensive experiments to evaluate how different variants of our sampling strategy affect the overall lane graph metrics (refer to Tables~\ref{tab:noise-ablation} and~\ref{tab:noise-levels-ablation}).
  
\end{itemize}

\begin{table}[H]
  \caption{Comparison 
 table between our method and the baseline approaches, evaluated using the GEO and TOPO metrics. For clarity, we omit the standard deviation for the methods that use diffusion models. Results marked with \(\dagger\) were taken from LaneExtraction~\cite{he2022lane}, where the same metrics and evaluation methodology are used. DM stands for Diffusion Model, and LRS refers to Lane Segmentation Refinement. Best results are highlighted in bold.}
  \label{tab:graph-metrics}
  \begin{tabularx}{\textwidth}{lcCCcCC}
    \toprule 
     \multirow{3}{*}{\textbf{Method}} & \multicolumn{3}{c}{\textbf{GEO Metrics}} & \multicolumn{3}{c}{\textbf{TOPO Metrics}} \\ \cmidrule{2-7}
                         & \textbf{F1 Score} & \textbf{Prec.} & \textbf{Rec.} & \textbf{F1 Score} & \textbf{Prec. }& \textbf{Rec.} \\
    \cmidrule{1-7}
    Standard U-Net \(\dagger\) & 0.786 & 0.811 & 0.762 & 0.747 & 0.622 & 0.679 \\
    LaneExtraction \(\dagger\)~\cite{he2022lane} & 0.828 & \textbf{0.835} & 0.821 & 0.748 & \textbf{0.774} & 0.724 \\
    Ensemble of DM~\cite{wu2022medsegdiff} & 0.658 & 0.746 & 0.589 & 0.576 & 0.670 & 0.506 \\
    LSR-DM (ours) & \textbf{0.841} &  0.833 & \textbf{0.849} & \textbf{0.774} & 0.759 & \textbf{0.789} \\
    
    \bottomrule
  \end{tabularx}
\end{table}

\begin{table}[H]
  \caption{Ablation 
 study on the components of our method: Cond. DDPM (conditional denoising diffusion probabilistic model \cite{ho2020denoising}) uses aerial patches, while Cond. DDIM (conditional denoising diffusion implicit model \cite{song2020denoising}) conditions the initial latent variable on the unrefined segmentation mask. The format of the results indicates the mean \(\pm\) std. across 10 repetitions for each experiment. Results highlighted in bold represent the best results.}
  \label{tab:components-ablation}
  \small
\begin{adjustwidth}{-\extralength}{0cm}
\begin{minipage}{\fulllength}
 \begin{tabularx}{\textwidth}{
  >{\centering\arraybackslash}p{1cm}
  >{\centering\arraybackslash}p{1cm}
  CCCCCC
 }
    \toprule 
     \multirow{2.5}{*}{\shortstack{\textbf{Cond.}\\\textbf{DDPM}}} & \multirow{2.5}{*}{\shortstack{\textbf{Cond.}\\\textbf{DDIM}}} & \multicolumn{3}{c}{\textbf{GEO Metrics}} & \multicolumn{3}{c}{\textbf{TOPO Metrics}} \\ \cmidrule{3-8}
                         & & \textbf{F1 Score} & \textbf{Prec. }& \textbf{Rec. }&\textbf{ F1 Score} & \textbf{Prec. }& \textbf{Rec.} \\
    \hline
    \xmark & \xmark  & \(0.010{~\pm~1 \times 10^{-3}}\) & \(0.050{~\pm~7 \times 10^{-3}}\) & \(0.005{~\pm~1 \times 10^{-3}}\) & \(0.001{~\pm~3 \times 10^{-4}}\) & \(0.012{~\pm~2 \times 10^{-3}}\) & \(0.0005{~\pm~2 \times 10^{-4}}\) \\
    
    \xmark & \cmark  & \(0.784{~\pm~3 \times 10^{-3}}\) & \(0.726{~\pm~4 \times 10^{-3}}\)  & \(\boldsymbol{0.852{~\pm~2 \times 10^{-3}}}\) & \(0.703{~\pm~4 \times 10^{-3}}\) & \(0.633{~\pm~5 \times 10^{-3}}\) & \(\boldsymbol{0.790{~\pm~4 \times 10^{-3}}}\) \\
    
    \cmark & \xmark & \(0.666{~\pm~4 \times 10^{-3}}\) & \(0.819{~\pm~4 \times 10^{-3}}\) & \(0.561{~\pm~3 \times 10^{-3}}\) & \(0.562{~\pm~5 \times 10^{-3}}\) & \(0.752{~\pm~6 \times 10^{-3}}\) & \(0.449{~\pm~5 \times 10^{-3}}\)  \\
    
    \cmark & \cmark & \(\boldsymbol{0.841{~\pm~2 \times 10^{-3}}}\) & \(\boldsymbol{0.833{~\pm~2 \times 10^{-3}}}\)  & \(0.849{~\pm~2 \times 10^{-3}}\) & \(\boldsymbol{0.774{~\pm~2 \times 10^{-3}}}\) & \(\boldsymbol{0.759{~\pm~2 \times 10^{-3}}}\) & \(0.789{~\pm~3 \times 10^{-3}}\) \\
    
    \bottomrule
  \end{tabularx}
\end{minipage}
\end{adjustwidth}
\end{table}

\begin{table}[H]
\caption{Ablation 
 study on adding Gaussian noise \(\epsilon \sim \mathcal{N}(\boldsymbol{0}, \boldsymbol{I})\) to the unrefined segmentation mask with different numbers of sampling steps \(S\). The format of the results indicates the mean \(\pm\) std. across 10 repetitions for each experiment. Results without added Gaussian noise lack standard deviation indicators, as the input remains consistent. Results highlighted in bold represent the best results.}
\label{tab:noise-ablation}

\small
\begin{adjustwidth}{-\extralength}{0cm}
\begin{minipage}{\fulllength}

\begin{tabularx}{\textwidth}{
  >{\centering\arraybackslash}p{1cm}
  >{\centering\arraybackslash}p{1cm}
  CCCCCC
 }
\toprule
\multirow{2}{*}{\boldmath \(S\) } & \multirow{2}{*}{ \textbf{Noise} } & \multicolumn{3}{c}{\textbf{GEO Metrics}} & \multicolumn{3}{c}{\textbf{TOPO Metrics}} \\ \cmidrule{3-8}
                         &                         & \textbf{F1 Score} & \textbf{Prec. }& \textbf{Rec. }&\textbf{ F1 Score} & \textbf{Prec.} & \textbf{Rec.} \\ \midrule
\multirow{2}{*}{10} & \checkmark & \(\boldsymbol{0.841{~\pm~2 \times 10^{-3}}}\)  & \(\boldsymbol{0.835{~\pm~2 \times 10^{-3}}}\) & \(0.847{~\pm~2 \times 10^{-3}}\) & \(\boldsymbol{0.774{~\pm~3 \times 10^{-3}}}\) & \(\boldsymbol{0.762{~\pm~3 \times 10^{-3}}}\) & \(0.786{~\pm~3 \times 10^{-3}}\) \\
                   & \xmark & 0.838 & 0.827 & \(\boldsymbol{0.848}\) & 0.772 & 0.755 & \(\boldsymbol{0.789}\) \\ \midrule

\multirow{2}{*}{25} & \checkmark & \(\boldsymbol{0.841{~\pm~2 \times 10^{-3}}}\) & \(\boldsymbol{0.833{~\pm~2 \times 10^{-3}}}\)  & \(0.849{~\pm~2 \times 10^{-3}}\) & \(\boldsymbol{0.774{~\pm~2 \times 10^{-3}}}\) & \(\boldsymbol{0.759{~\pm~2 \times 10^{-3}}}\) & \(0.789{~\pm~3 \times 10^{-3}}\) \\
                   & \xmark & 0.837 & 0.825 & \(\boldsymbol{0.850}\) & 0.772 & 0.753 & \(\boldsymbol{0.792}\) \\ \midrule
                   
\multirow{2}{*}{50} & \checkmark & \(\boldsymbol{0.840{~\pm~2 \times 10^{-3}}}\) & \(\boldsymbol{0.831{~\pm~2 \times 10^{-3}}}\) & \(0.849{~\pm~2 \times 10^{-3}}\) & \(0.773{~\pm~3 \times 10^{-3}}\) & \(\boldsymbol{0.756{~\pm~3 \times 10^{-3}}}\) & \(0.790{~\pm~2 \times 10^{-3}}\) \\
                   & \xmark &0.838 & 0.825 & \(\boldsymbol{0.852}\) & \textbf{0.775} & \(\boldsymbol{0.756}\) & \(\boldsymbol{0.795}\) \\ \midrule
                   
\multirow{2}{*}{100} & \checkmark & \(\boldsymbol{0.840{~\pm~3 \times 10^{-3}}}\) & \(\boldsymbol{0.831{~\pm~4 \times 10^{-3}}}\) & \(0.849{~\pm~3 \times 10^{-3}}\) & \(\boldsymbol{0.773{~\pm~4 \times 10^{-3}}}\) & \(\boldsymbol{0.756{~\pm~4 \times 10^{-3}}}\) & \(0.790{~\pm~3 \times 10^{-3}}\) \\
                   & \xmark & 0.838 & 0.825 & \(\boldsymbol{0.852}\) & 0.772 & 0.752 & \(\boldsymbol{0.793}\) \\ \midrule
                   
\multirow{2}{*}{250} & \checkmark & \(\boldsymbol{0.839{~\pm~1 \times 10^{-3}}}\) & \(\boldsymbol{0.829{~\pm~2 \times 10^{-3}}}\) & \(0.849{~\pm~1 \times 10^{-3}}\) & \(0.772{~\pm~2 \times 10^{-3}}\) & \(\boldsymbol{0.755{~\pm~2 \times 10^{-3}}}\) & \(0.790{~\pm~1 \times 10^{-3}}\) \\
                   & \xmark & 0.838 & 0.825 & \(\boldsymbol{0.852}\) & \textbf{0.773} & 0.752 & \(\boldsymbol{0.795}\) \\ \midrule
                   
\multirow{2}{*}{500} & \checkmark & \(\boldsymbol{0.841{~\pm~1 \times 10^{-3}}}\) & \(\boldsymbol{0.831{~\pm~2 \times 10^{-3}}}\) & \(0.851{~\pm~1 \times 10^{-3}}\) & \(\boldsymbol{0.773{~\pm~2 \times 10^{-3}}}\) & \(\boldsymbol{0.756{~\pm~2 \times 10^{-3}}}\) & \(0.790{~\pm~2 \times 10^{-3}}\) \\
                   & \xmark & 0.838 & 0.824 & \(\boldsymbol{0.852}\) & \textbf{0.773} & 0.752 & \(\boldsymbol{0.795}\) \\ \midrule
                   
\multirow{2}{*}{1000} & \checkmark &\(\boldsymbol{0.838{~\pm~1 \times 10^{-3}}}\) & \(\boldsymbol{0.828{~\pm~1 \times 10^{-3}}}\) & \(0.849{~\pm~1 \times 10^{-3}}\) & \(\boldsymbol{0.771{~\pm~1 \times 10^{-3}}}\) & \(\boldsymbol{0.753{~\pm~1 \times 10^{-3}}}\) & \(0.789{~\pm~2 \times 10^{-3}}\) \\
                   & \xmark & \textbf{0.838} & 0.824 & \(\boldsymbol{0.852}\) & \textbf{0.771} & 0.751 & \(\boldsymbol{0.792}\) \\ \bottomrule
\end{tabularx}

\end{minipage}
\end{adjustwidth}
\end{table}

\begin{table}[H]

\caption{Ablation 
study on different noise levels added to the unrefined mask, according to the forward process used during training, with sampling steps fixed at \(S = 25\). FS refers to the number of forward diffusion steps applied to the unrefined segmentation mask. The format of the results indicates the mean \(\pm\) std. across 10 repetitions for each experiment. Results highlighted in bold represent the best results. Results underlined denote the worst, occurring when the starting latent variable \(x_T\) consists purely of Gaussian noise.
}
\label{tab:noise-levels-ablation}
\small

\begin{adjustwidth}{-\extralength}{0cm}
\begin{minipage}{\fulllength}
\begin{tabularx}{\textwidth}{CCCCCCC}
\toprule
\multirow{2.5}{*}{ \textbf{Noise Level }} & \multicolumn{3}{c}{\textbf{GEO Metrics}} & \multicolumn{3}{c}{\textbf{TOPO Metrics}} \\ \cmidrule{2-7}
                         & \textbf{F1 Score} & \textbf{Prec.} & \textbf{Rec. }& \textbf{F1 Score }& \textbf{Prec. }& \textbf{Rec.} \\ \midrule
0\% (0 FS) & \(0.837{~\pm~0  \times 10^{-3}}\) & \(0.825{~\pm~0 \times 10^{-3}}\)& \(0.850{~\pm~0 \times 10^{-3}}\)& \(0.772{~\pm~0 \times 10^{-3}}\)& \(0.753{~\pm~0 \times 10^{-3}}\)&\(0.792{~\pm~0 \times 10^{-3}}\)\\

10\% (100 FS) & \(0.838{~\pm~1 \times 10^{-3}}\)& \(0.826{~\pm~1 \times 10^{-3}}\)& \(\boldsymbol{0.851{~\pm~2 \times 10^{-3}}}\)& \(0.773{~\pm~2 \times 10^{-3}}\)& \(0.754{~\pm~2 \times 10^{-3}}\)& \(\boldsymbol{0.793{~\pm~2 \times 10^{-3}}}\)\\

25\% (250 FS) & \(0.837{~\pm~1 \times 10^{-3}}\)& \(0.825{~\pm~1 \times 10^{-3}}\)& \(0.850{~\pm~1 \times 10^{-3}}\)& \(0.772{~\pm~2 \times 10^{-3}}\)& \(0.753{~\pm~2 \times 10^{-3}}\)& \(0.792{~\pm~2 \times 10^{-3}}\)\\

50\% (500 FS) & \(\boldsymbol{0.839{~\pm~3 \times 10^{-3}}}\)& \(0.832{~\pm~3 \times 10^{-3}}\)& \(0.846{~\pm~3 \times 10^{-3}}\)& \(\boldsymbol{0.775{~\pm~3 \times 10^{-3}}}\)& \(0.762{~\pm~3 \times 10^{-3}}\)& \(0.787{~\pm~0 \times 10^{-3}}\)\\

75\% (750 FS) & \(0.826{~\pm~2 \times 10^{-3}}\)& \(\boldsymbol{0.842{~\pm~2 \times 10^{-3}}}\)& \(0.811{~\pm~3 \times 10^{-3}}\)& \(0.757{~\pm~2 \times 10^{-3}}\)& \(\boldsymbol{0.774{~\pm~3 \times 10^{-3}}}\)& \(0.741{~\pm~3 \times 10^{-3}}\)\\

90\% (900 FS) & \(0.781{~\pm~4 \times 10^{-3}}\)& \(0.839{~\pm~5 \times 10^{-3}}\)& \(0.730{~\pm~4 \times 10^{-3}}\)& \(0.697{~\pm~5 \times 10^{-3}}\)& \(0.774{~\pm~5 \times 10^{-3}}\)& \(0.634{~\pm~5 \times 10^{-3}}\)\\

100\% (1000 FS) & \(\underline{0.666{~\pm~4 \times 10^{-3}}}\)& \(\underline{0.819{~\pm~4 \times 10^{-3}}}\)& \(\underline{0.561{~\pm~3 \times 10^{-3}}}\)& \(\underline{0.562{~\pm~5 \times 10^{-3}}}\)& \(\underline{0.752{~\pm~6 \times 10^{-3}}}\)& \(\underline{0.449{~\pm~5 \times 10^{-3}}}\)  \\ 

\bottomrule
\end{tabularx}
\end{minipage}
\end{adjustwidth}
\end{table}

In the following sections, we present the related work relevant to our method, which includes (1) lane graph extraction from onboard sensors, (2) lane graph extraction from aerial imagery, and 3) the use of diffusion models for segmentation tasks.

\subsection{Lane Graph Extraction from Onboard Sensors}

Several previous works~\cite{homayounfar2018hierarchical, zurn2021lane, homayounfar2019dagmapper, zhou2021automatic, zhang2022hierarchical} have utilized either street-view images or 3D point clouds collected from onboard sensors such as cameras and LiDAR to extract lane graphs. Street-view images provide rich semantic and visual cues, enhancing the detection of lane centerlines and lane-level topology. Furthermore, 3D point clouds offer precise geometric and elevation data, capturing lane shapes and boundaries. Inspired by human annotators, Homayounfar et al.~\cite{homayounfar2018hierarchical} train a hierarchical recurrent neural network to sequentially identify and trace lane boundaries, first attending to initial boundary regions and then outputting complete structured polylines. Zürn et al.~\cite{zurn2021lane} employ a multi-modal bird’s-eye view (BEV) representation that integrates LiDAR, RGB, vehicle segmentation masks, and a semantic map (excluding vehicles) as input to train a graph region-based convolutional neural network (Graph R-CNN)~\cite{yang2018graph}. The network is trained with a regression loss and the method uses a post-processing algorithm to predict the lane graph within local BEV patches. The post-processing step eliminates false positive connections by applying probability and distance thresholds on the output lane graph. In both works, the ground truth data are obtained by projecting 3D LiDAR point clouds into a BEV representation. DAGMapper~\cite{homayounfar2019dagmapper} uses LiDAR intensity images as input to a recurrent convolutional network with three heads that parameterize a directed acyclic graphical model (DAGM), which iteratively constructs the lane graph. Formulating the problem as a directed acyclic graph, i.e., a graph without cycles, is advantageous, as it mirrors the approach human annotators would take when labeling a lane graph on an aerial image. Zhang et al.~\cite{zhang2022hierarchical} combine sensing data from LiDAR and cameras to train a  hierarchical fully convolutional network that predicts a lane-level road network. Combining LiDAR and camera data leverages the strengths of both modalities, resulting in a more robust and reliable lane graph. Zhou et al. employ an encoder--decoder architecture to generate semantic maps from camera and LiDAR data, accumulating frames into a LiDAR-synchronized bird’s-eye view (BEV) projection. They also incorporate OpenStreetMap (OSM) data to help the model infer lane connections at intersections. Can et al.~\cite{can2021structured, can2022topology} employ images of a single onboard camera. In their first work~\cite{can2021structured}, they use a Transformer~\cite{vaswani2017attention} with two heads that predicts lane center-lines and bounding boxes of objects. A Transformer is used instead of a CNN to more effectively model long-range dependencies and global context which might be beneficial for extracting the lane-level graph. In their second work~\cite{can2022topology}, they refine their first approach by replacing the object detection head with a minimal cycle head. This head is designed to identify the smallest cycles formed by directed curve segments between intersections.

Although extracting the lane graph using an ego-vehicle equipped with onboard sensors typically results in highly quality lane-level graphs, this approach has several drawbacks. The main disadvantages include high costs and extended data collection times, making it difficult to scale to large geographical areas. Furthermore, onboard sensors have limited fields of view and struggle with occlusions caused by other vehicles, buildings, or vegetation, leading to incomplete or inconsistent lane graphs. Additionally, maintaining and operating sensor-equipped vehicles requires logistical resources and infrastructure, further hindering scalability. 

{In contrast, aerial imagery offers broader coverage with faster data collection times and does not depend on vehicle-based infrastructure, making it a more scalable and cost-effective solution. Its consistent top--down perspective simplifies lane detection, provides a wider field of view that reduces occlusion issues, and enables mapping of remote or hard-to-reach areas that may be inaccessible or impractical for sensor-equipped vehicles.}

\subsection{Lane Graph Extraction from Aerial Imagery}

{These advantages position aerial imagery as a compelling alternative to onboard sensors for extracting lane-level graphs.} However, aerial imagery also poses challenges, including low ground object resolution, occlusions caused by urban infrastructure and vehicles, as well as variations in road texture and lighting conditions. Another significant challenge is the scarcity of datasets for this task, though recent releases~\cite{he2022lane, buchner2023learning} have enabled the development of some methods~\cite{he2022lane, buchner2023learning, blayney2024bezier} to address this task.

Büchner et al.~\cite{buchner2023learning} employ centerline regressor models with a virtual agent positioned at a specific point in a BEV patch to iteratively predict the successor sub-graph relative to the agent’s current location. They identify potential nodes within corridors likely to contain the successor sub-graph, which is then predicted using a causal variant of a message-passing network~\cite{braso2020learning}. The causality prior enables the network to incorporate information about predecessor and successor features during message passing, allowing it to model directional relationships between nodes. Blayney et al.~\cite{blayney2024bezier} represent the lane graph as a Bézier Graph, where nodes define the start and end points of cubic Bézier curves, parameterized by four control points, and edges represent the curves connecting these points. Nodes store the position of the ending points and direction vectors, while edges contain two length values. The intermediate control points are computed from the attributes of the nodes and edges. In this graph, nodes aligned in a straight line are significantly reduced. This compact representation decreases the total number of nodes, enabling the use of a Transformer \cite{vaswani2017attention} architecture. An encoder Transformer processes visual features extracted from the aerial image by a backbone CNN. A Transformer decoder then performs cross-attention between the encoder output and a set of node and edge queries, producing node and edge embeddings. These embeddings are subsequently passed through multilayer perceptron (MLP) heads to predict the parameters of the nodes and edges of the Bézier Graph. These methods~\cite{buchner2023learning, blayney2024bezier} can be considered graph-based approaches, as they do not produce lane segmentation masks as the final output of the learning process, although they may utilize segmentation masks within their training pipelines. As mentioned earlier, graph-based methods typically yield high connectivity and completeness in lane graphs. However, they require algorithms to aggregate the local graphs predicted within each patch, making them slower than segmentation-based approaches.

In contrast to graph-based methods, segmentation-based approaches offer faster inference times and simpler solutions for merging different segmented aerial patches. Once the entire area is merged, the lane graph is extracted using a conventional segmentation-to-graph algorithm. He et al.~\cite{he2022lane} tackle the problem this way, where the final outputs of the learning process are lane segmentation masks. These masks are subsequently post-processed using a traditional segmentation-to-graph algorithm. This approach divides the task into two subtasks: lane extraction in non-intersection areas and turning lane extraction at intersections. This division simplifies the learning process for the CNNs, as the visual patterns and lanes differ significantly between intersection and non-intersection areas. For instance, lanes in non-intersection areas do not cross, while those at intersections often do, and intersection areas occupy only a small portion of the aerial image compared to non-intersection areas. Due to the complex visual patterns and intricate lane connections, intersection areas may also require additional pre- or post-processing. In the first subtask, a D-LinkNet~\cite{zhou2018d}, a CNN based on a U-Net~\cite{ronneberger2015u} with dilated convolutions~\cite{yu2015multi, chen2018encoder}, is employed to extract segmentation and direction maps of the lanes. Direction maps are segmentation maps where colors encode the driving direction of the lanes. Using direction maps alongside lane segmentation masks has been shown to improve performance in road segmentation, a task closely related to lane segmentation. The direction map is then used to predict the direction of the graph. In the second subtask, a classifier distinguishes between valid and invalid lane connections. It leverages aerial images, potential segmentation masks of the lane connections, and positional information from the terminal nodes (defined as nodes with exactly one incoming or outgoing edge). 
These inputs provide geometric cues and visual features that help the classifier differentiate between valid and invalid connections. The potential connection segmentation masks are generated using a network similar to the one employed in the first subtask. Lane connections are selected from a pool of ordered terminal node pairs. Candidate terminal node pairs are constructed based on a distance threshold between terminal nodes within the lane graph derived from non-intersection areas extracted in the first subtask. This approach is effective because terminal nodes from different crossroads are typically much farther apart than those from the same crossroad, reflecting the fact that most crossroads are not situated close to one another. This process is followed by predicting segmentation masks for valid connections, which may represent either curves (for turns) or straight lines, using a D-LinkNet similar to the one used previously. Once the segmentation mask is obtained, the segmentation-to-graph algorithm is applied to extract the subgraph of the lane connections. Finally, the lane graphs extracted from both subtasks are integrated by using the terminal nodes as shared junctions.

Our method builds upon this previous work but is specifically tailored to improve lane extraction in non-intersection areas (first subtask). It addresses a key limitation of segmentation-based approaches: incomplete and noisy lane masks. By producing sharp and complete lane masks, the subsequent segmentation-to-graph algorithm yields higher-quality lane graphs, specially in terms of connectivity. We leave the second subtask for future work, as it presents additional challenges, including applying diffusion models on intersection areas and integrating geo-positional information for the terminal nodes. {Nevertheless, we believe our method could be extended to address these challenges with further modifications. We present several potential approaches in Section~\ref{sec:discussion}.}

\subsection{Diffusion Models for Image Segmentation}

\label{sec:diffusion-models-for-image-segmentation}

Diffusion models (DMs) are generative models that synthesize new data from Gaussian noise. The intuition behind these types of models is to corrupt a clean sample by progressively adding noise and then learn to reverse this process using a random process. Thanks to the stochasticity of the learned reverse process, new data can be generated. Diffusion models comprise three principal components: the  forward process 
, the denoising process, and the sampling procedure (also referred to as inference)~\cite{chang2023design}. During the forward process, noise is progressively added to the data over a fixed time horizon. This process can be formalized as a Markov chain, i.e., each state depends only on the previous state, with Gaussian transition distributions following a variance schedule that controls the balance between signal and noise. The denoising process progressively removes the noise introduced during the forward process. While it can also be modeled as a Markov chain, computing the true transition distributions is intractable, so these distributions are approximated using parameterized models. To learn the parameters, a neural network is trained by optimizing the lower bound loss. Once trained, the neural network is used in the sampling procedure to generate new data from noise sampled from a standard normal distribution. Denoising diffusion probabilistic models (DDPMs)~\cite{ho2020denoising} provide a simple parameterization for learning the denoising process, resulting in the generation of high-quality images, which subsequent works~\cite{dhariwal2021diffusion, nichol2021improved} further improve. Denoising diffusion implicit models (DDIMs)~\cite{song2020denoising} accelerate the sampling procedure by generalizing DDPMs to non-Markovian chains, substantially reducing the number of sampling steps relative to the forward steps.

Conditional diffusion models are diffusion models that condition data generation on additional inputs such as class labels~\cite{dhariwal2021diffusion, nichol2021improved} or embeddings~\cite{rombach2022high} derived from external inputs, including text, semantic maps, or images. These conditional inputs guide the model to produce more accurate and relevant outputs. Conditional diffusion models are employed in image segmentation tasks due to their flexibility in creating ensembles through different random seeds during inference, which makes the segmentation output more robust, and their ability to accurately segment complex visual patterns, areas where CNNs often struggle. However, these models also have some drawbacks, including higher inference times and challenges in achieving deterministic segmentation outputs. Despite these limitations, conditional diffusion models have demonstrated state-of-the-art performance for image segmentation in specific domains~\cite{amit2021segdiff, wolleb2022diffusion, wu2022medsegdiff, wu2023medsegdiff}. SegDiff~\cite{amit2021segdiff} utilizes a conditional diffusion model for image segmentation across diverse domains, where segmentation masks are diffused, conditioned on the corresponding images. This method generates an ensemble of segmentation masks through multiple sampling processes, which are then averaged to produce the final segmentation mask. Building on SegDiff, MedSegDiff~\cite{wu2022medsegdiff} enhances the original architecture of SegDiff to segment medical imagery by introducing a feature frequency parser (FF-Parser), an attention module that operates on Fourier-space of the image features. This module can be viewed as a learnable counterpart of frequency filters widely used in digital image processing. MedSegDiff-V2~\cite{wu2023medsegdiff} further refines this approach by employing a Transformer architecture within the FF-Parser module, achieving improved segmentation performance compared to its predecessor.

Although diffusion models have been successfully employed for image segmentation, ensembles of diffusion models struggle with thin structures, such as lane segmentation masks. This limitation arises from slight lateral variations in the segmented pixels across the ensemble, resulting in a blurred average segmentation mask (see the second row of Figures~\ref{fig:results-tile-A} and~\ref{fig:results-tile-B}), which degrades the quality of the final lane graph. By contrast, this issue is less pronounced for larger objects, where minor discrepancies along the contours have minimal impact on the overall segmentation mask. Our method addresses this challenge by using conditional diffusion models for segmenting the lane masks without relying on an ensemble. Instead, we initiate the sampling procedure from a latent variable conditioned on the output of a CNN and follow a deterministic sampling process (see Section~\ref{sec:segmentation-refinement-stage}).


\section{{Materials and Methods}}
\label{sec:materials-and-method}

\begin{figure}[H]
  \includegraphics[width=.99\textwidth]{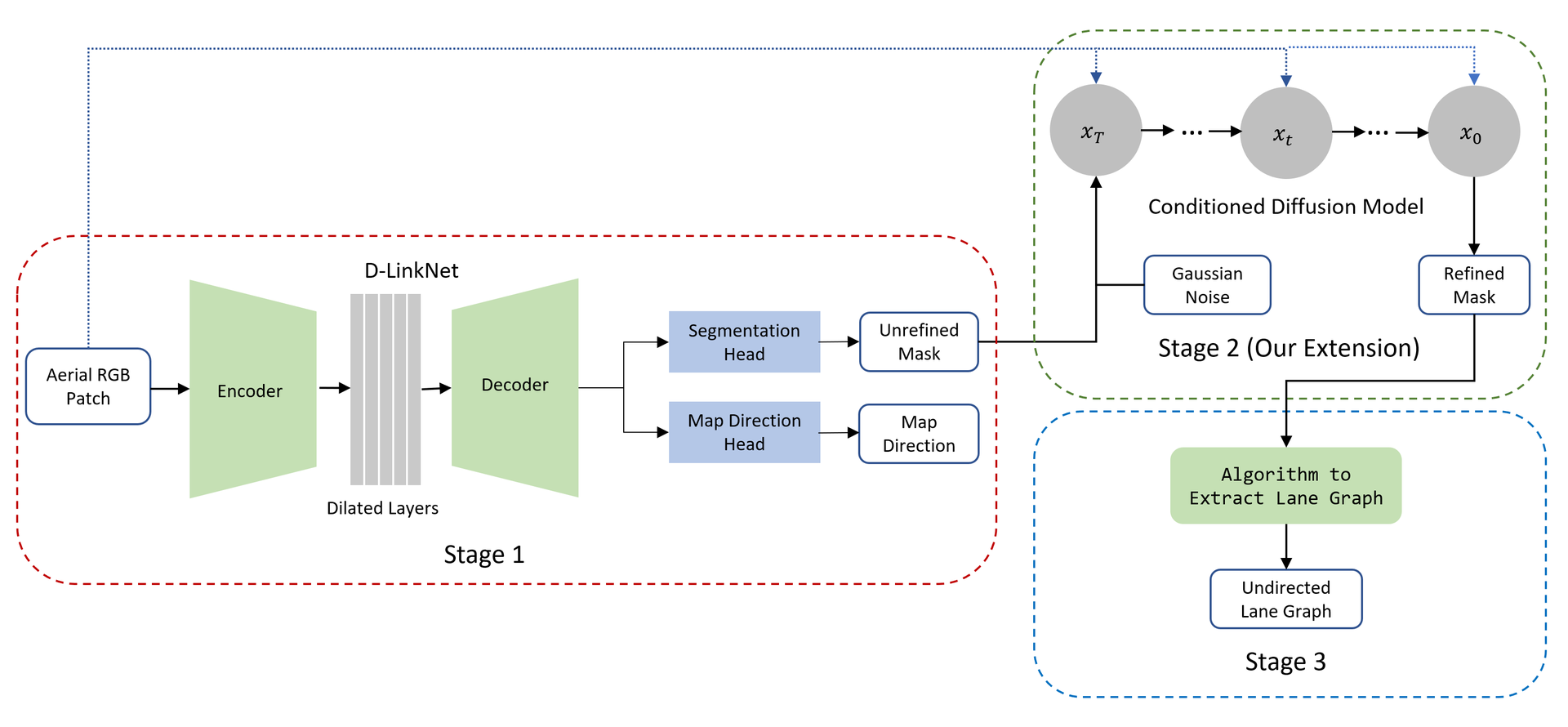}
  \caption{Overall pipeline of our method. During inference, the aerial RGB patch is first fed into the D-LinkNet~\cite{zhou2018d}, which outputs an unrefined segmentation mask and an unrefined direction map. Gaussian noise is then added to the unrefined segmentation mask to create the starting latent variable \(x_T\) (starting point instead of Gaussian noise) for the DDIM~\cite{song2020denoising} sampling. After several sampling steps, a refined segmentation mask \(x_0\)  is generated. In Stage 2, the blue dotted arrows indicate conditioning, while the solid black arrow represents the input variables for the diffusion model. Finally, the refined segmentation mask is passed to the segmentation-to-graph algorithm to produce the final lane graph.}
  \label{fig:overall-pipeline}
\end{figure}

Our method is divided into three distinct stages: (1) lane segmentation, (2) lane segmentation refinement, and (3) lane graph extraction (see Figure~\ref{fig:overall-pipeline}). Each stage addresses a specific challenge: the First Stage utilizes a CNN to perform an initial segmentation of the lanes; the Second Stage employs a conditional diffusion model and a novel sampling strategy to refine the lane segmentation masks; and the Third Stage converts these refined masks into a graph using a conventional segmentation-to-graph algorithm. It is important to note that only the first two stages involve training and inference phases, whereas the third stage requires no training, relying instead on a traditional rule-based algorithm.

Our primary contribution lies in the lane segmentation refinement stage, while the other two stages are taken from LaneExtraction~\cite{he2016deep}. As a result of this refinement process, the lane masks appear significantly sharper, i.e., lane masks exhibit clearly delineated boundaries, and lane segments become continuous (see Figures~\ref{fig:results-tile-A} and~\ref{fig:results-tile-B} for visual examples). This improvement in the lane segmentation masks leads to a higher-quality lane graph after the extraction process in the Third Stage (see quantitative evaluations in \mbox{Table~\ref{tab:graph-metrics}}). In addition to enhancing the quality of the resulting lane graph, our approach does not require sequential training, since the models for stages one and two can be trained independently. This independence allows for the flexible use of different architectures and training procedures for the segmentation and diffusion models, enabling parallel training. This parallelization improves training efficiency but requires more computational resources compared to methods that only require one model (refer to Table~\ref{tab:computational-resources}). However, at inference time, the sequential order of the stages must be maintained (see Figure~\ref{fig:overall-pipeline}).

\label{fig:diffusion-architecture}

\begin{table}[H]
  \caption{{Computational resources used by our method and the baselines. We evaluate three resource metrics: training time, inference time, and model size. Both training and inference for all methods were conducted on a single NVIDIA GeForce RTX 3090 GPU.}}
  \label{tab:computational-resources}

\begin{tabularx}{\textwidth}{lCCC}
    \toprule 
    \multirow{2}{*}{\textbf{Method}} & \multicolumn{3}{c}{\textbf{Computational Resources}} \\ \cmidrule{2-4}
                            & \textbf{Training Time} &\textbf{ Inference Time} & \textbf{Model Size (MB) }\\
    \cmidrule{1-4}
    Standard U-Net  & 7.3 h & 445 s & 119 \\
    LaneExtraction~\cite{he2022lane} & 7.2 h & 230 s & 142 \\
    Ensemble of DM~\cite{wu2022medsegdiff} & 37.3 h & 13,410 s & 222 \\
    LSR-DM (ours) & 80.9 h & 714 s & 305 \\
    \bottomrule
\end{tabularx}

\end{table}

For the lane segmentation refinement, we employ a conditional diffusion model due to its effectiveness in addressing complex visual challenges that CNNs typically struggle with. The diffusion model refines the lane masks by reducing false positives and false negatives, thereby enhancing both their completeness and sharpness. We train the diffusion model conditioned on the RGB aerial images, following the framework of Improved DDPM~\cite{nichol2021improved}. For further technical details, we refer the reader to Section~\ref{sec:segmentation-refinement-stage}.

\begin{figure}[H]
  \includegraphics[width=.95\textwidth]{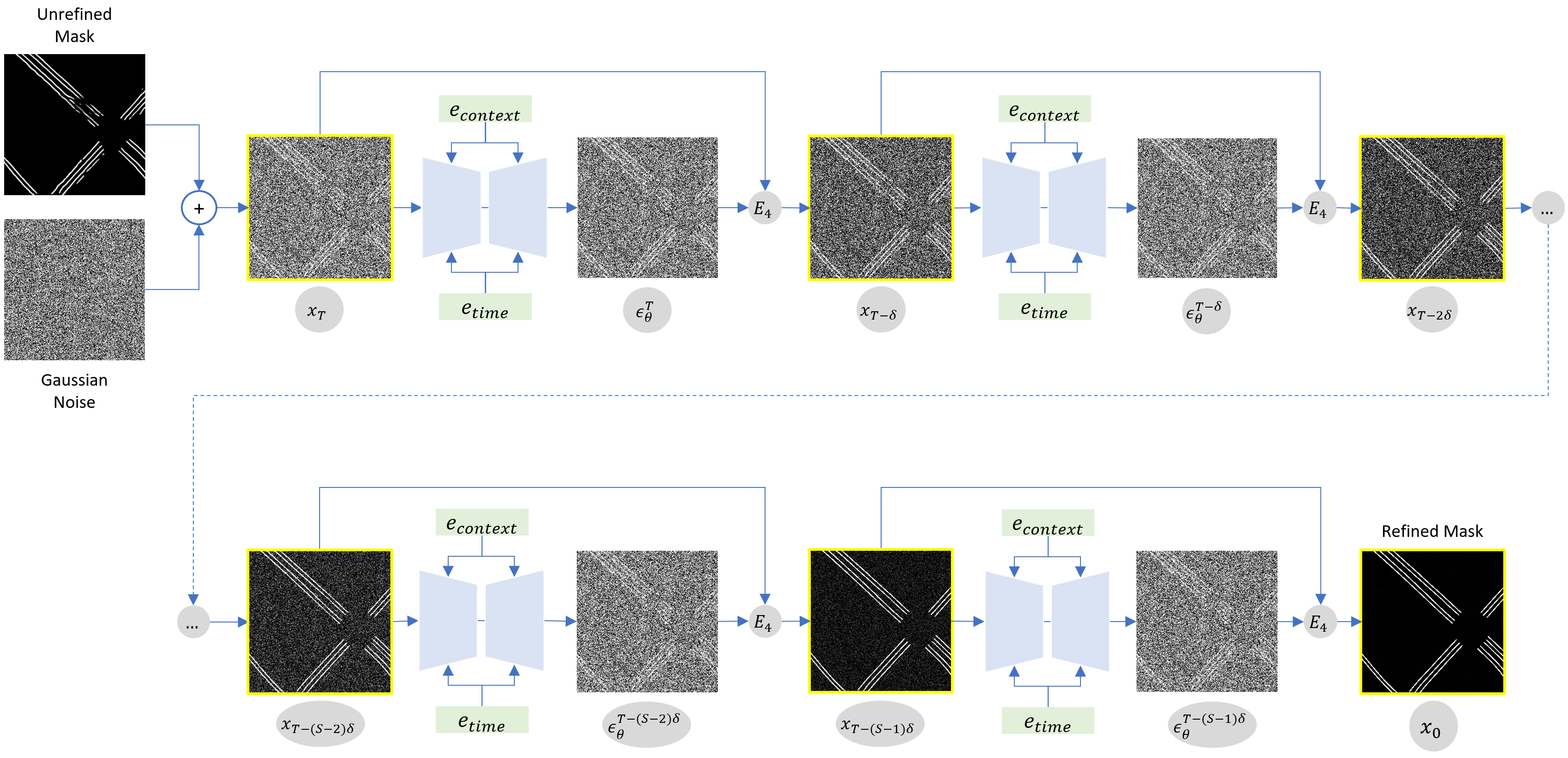}
  \caption{Conditional DDIM~\cite{song2020denoising} sampling process. Gaussian noise is added to the unrefined segmentation mask (from Stage 1) to generate the initial latent variable \(x_T\). Then, several DDIM sampling steps, as described in Equation~\eqref{eq:x-t-delta-equation}, are applied to progressively refine the segmentation mask, resulting in the final output \(x_0\).}
  \label{fig:diffusion-sampling}
\end{figure}

For sampling, we employ DDIM~\cite{song2020denoising}, which improves efficiency and enables a deterministic sampling path. Unlike standard diffusion models that start DDIM sampling from Gaussian noise, our approach conditions the initial latent variable on the output of the CNN (referred to as the unrefined segmentation mask), which is then refined through a few DDIM sampling steps (see Figure~\ref{fig:diffusion-sampling} for a visualization of the process). The unrefined segmentation mask serves a good approximation of the ground truth segmentation mask, thereby reducing the number of plausible sampling trajectories for the diffusion model. In contrast, starting from Gaussian noise makes the process significantly more difficult due to the vastly larger space of potential sampling trajectories. This observation is supported by experimental results in Table~\ref{tab:noise-levels-ablation}, where starting from pure Gaussian noise (last row) leads to worse performance compared to initializing from a latent variable conditioned on the unrefined segmentation mask (other rows of Table~\ref{tab:noise-levels-ablation}).

We explore the conditioning of the starting latent variable with various strategies, including directly starting from the unrefined segmentation mask, adding Gaussian noise to it, and applying noise through several forward diffusion steps (as during training). The rationale behind adding Gaussian noise is to make the latent variable more closely resemble a sample from a Gaussian distribution, which is the expected input for DDIM sampling. Applying noise via forward steps better aligns with the noisy samples used during training for learning the denoising process, but we also include a comparison with the simpler approach of adding Gaussian noise directly. We show experimentally that adding noise to the unrefined segmentation mask, either directly or through the forward steps, enhances the performance metrics and stability of the model compared to directly using the unrefined mask in DDIM sampling (see Tables~\ref{tab:noise-ablation} and~\ref{tab:noise-levels-ablation}). We present our analysis of the results in  Section~\ref{sec:quantitative-results}.

In the following sections, we provide detailed technical explanations of the three stages of our method, as well as relevant implementation details.

\subsection{First Stage: Lane Segmentation}

For the First Sage, we use the same architecture as LaneExtraction~\cite{he2022lane}, which employs a D-LinkNet~\cite{zhou2018d} with two heads: one outputs the segmentation mask while the other predicts the lane direction map. D-LinkNet includes dilated convolutional layers, which are beneficial for lane segmentation from aerial imagery because they expand the receptive field without increasing the number of parameters or reducing spatial resolution. This allows the network to capture broader contextual information essential for identifying continuous structures like lanes while preserving fine details such as lane edges and markings. Joint training of lane segmentation masks with direction maps has proven effective in improving the segmentation performance of both road-~\cite{batra2019improved} and lane-level~\cite{he2022lane} networks. We also follow a similar preprocessing procedure, which involves extracting random patches of fixed size \( N_s \times N_s \) (\( N_s = 1024 \)) from the training tiles (see Section~\ref{sec:dataset} for more details on the dataset), where the training tiles have a size of \( 2048 \times 2048 \). This patch size represents a sweet spot, providing sufficient context while ensuring a diverse variety of patches. After selecting a random patch, we apply augmentation techniques, including random rotations and random adjustments to color and brightness, to improve the robustness of the model to lighting and orientation variations, which enhances generalization. We let \( \mathcal{T} = \{(P_i, s_i, d_i)\}_{i=1}^{n}\) be the patch training set of \(n\) triplets induced by the preprocessing procedure, where \(P_i\) is an aerial training patch, \(s_i\) is its ground truth (GT) segmentation mask, and \(d_i\) is its ground truth direction map. The segmentation loss \(\mathcal{L}_{seg}\) is a combination of the cross-entropy loss and the dice loss  for the  segmentation head and the \( \mathcal{L}_2 \) loss for the direction map head is formally defined as follows:

\begin{equation}
  \mathcal{L}_{seg} =  \mathcal{L}_2(d_i, \hat{d}_i) + \frac{1}{2}(\mathcal{L}_{ce}(s_i,\hat{s}_i) + \mathcal{L}_{dice}(s_i, \hat{s}_i)),
  \label{eq:segmentation-loss}
\end{equation}
where \( \hat{s}_i \) and \( \hat{d}_i \) are the predictions for the segmentation and direction map, respectively, output by the CNN given the aerial patch \(P_i\). 
{The CNN performs multitask learning, simultaneously predicting lane direction maps and segmentation masks. Both tasks are considered equally important. The loss function in Equation~\eqref{eq:segmentation-loss} consists of three components: the \( \mathcal{L}_2 \) loss encourages the model to accurately estimate direction maps, while the cross-entropy (CE) loss and Dice loss guide the model to obtain correct segmentation masks. This dual segmentation loss formulation leverages the strengths of both approaches: CE promotes pixel-wise classification accuracy while Dice loss mitigates class imbalance, which is particularly significant in lane segmentation masks. The chosen weights, \(1.0\) for the \( \mathcal{L}_2 \) loss and \(0.5\) for both CE and Dice losses, which together sum to \(1.0\) for the segmentation branch, ensure that direction and segmentation contribute equally to the overall loss. This weighting strategy promotes balanced optimization between the two tasks, without favoring one over the other.}

\begin{figure}[H]
  \includegraphics[width=.95\textwidth]{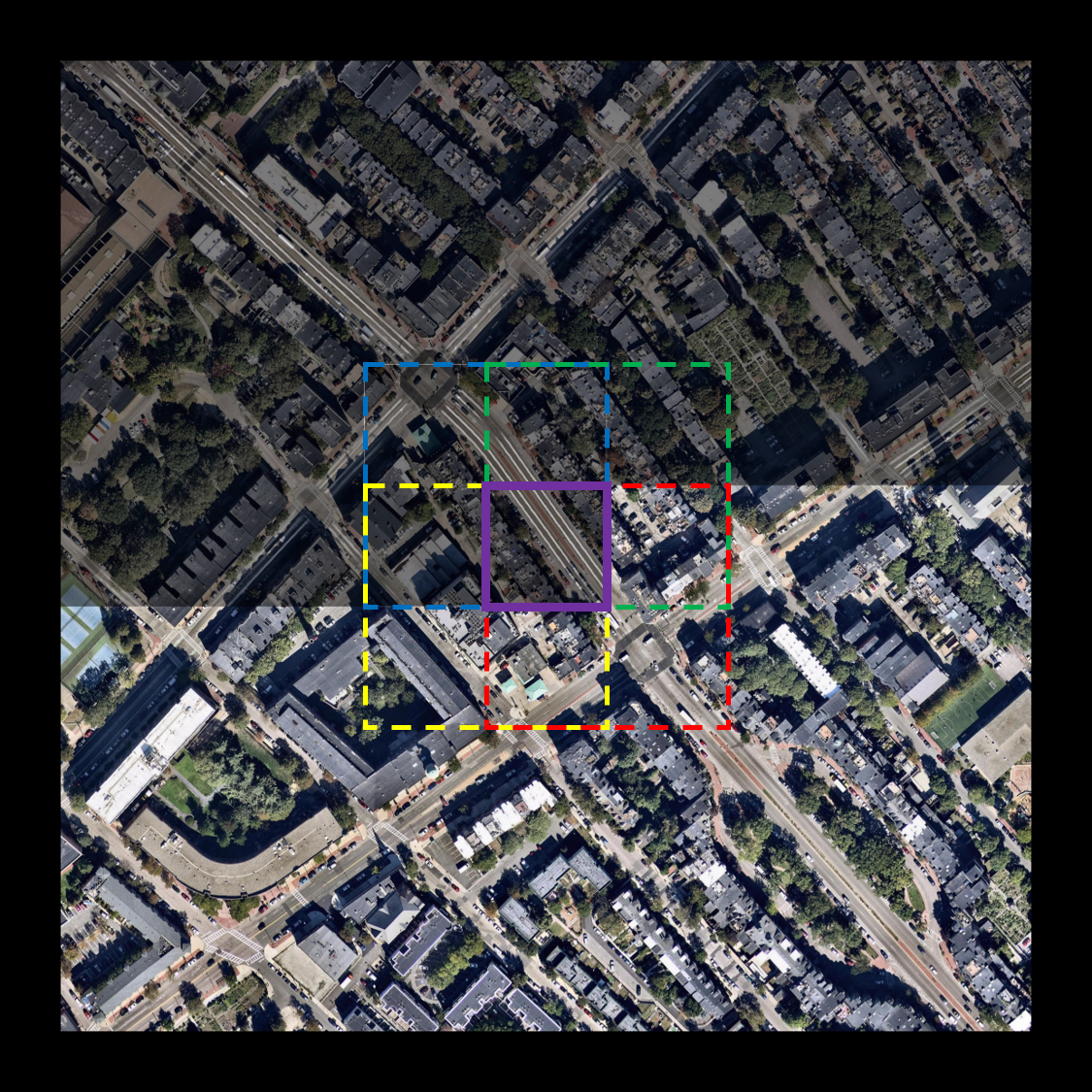}
  \caption{Sliding windows during inference. The shaded area represents the portion of the image already predicted. The dotted squares, each in a different color, illustrate the sliding windows. The small purple-framed region indicates the overlap between the four sliding windows, where the final output is obtained by averaging the results from these windows. This inference process is applied in both Stage 1 and Stage 2, with Stage 2 also incorporating the unrefined segmentation mask as an additional input.}
  \label{fig:sliding-window-prediction}
\end{figure}

The inference is performed using a sliding window, with the same size as the training patches (\(1024 \times 1024\)), applied to the full test image tiles of size \(4096 \times 4096\) with both horizontal and vertical strides of \(512\). For each sliding window, a prediction is made using the trained network. Once the entire tile is covered, predictions in overlapping areas (see dashed squares in different colors in Figure~\ref{fig:sliding-window-prediction}) are averaged to produce the final output. The average is taken over four overlapping regions, which enhances robustness. As we only need to apply the sliding window \(7\times7 = 49\) times to cover the entire tile, efficiency is also guaranteed.

\subsection{Second Stage: Lane Segmentation Refinement}
\label{sec:segmentation-refinement-stage}

\begin{figure}[H]
  \subfloat[ ]{
  \includegraphics[clip,width=.95\textwidth]{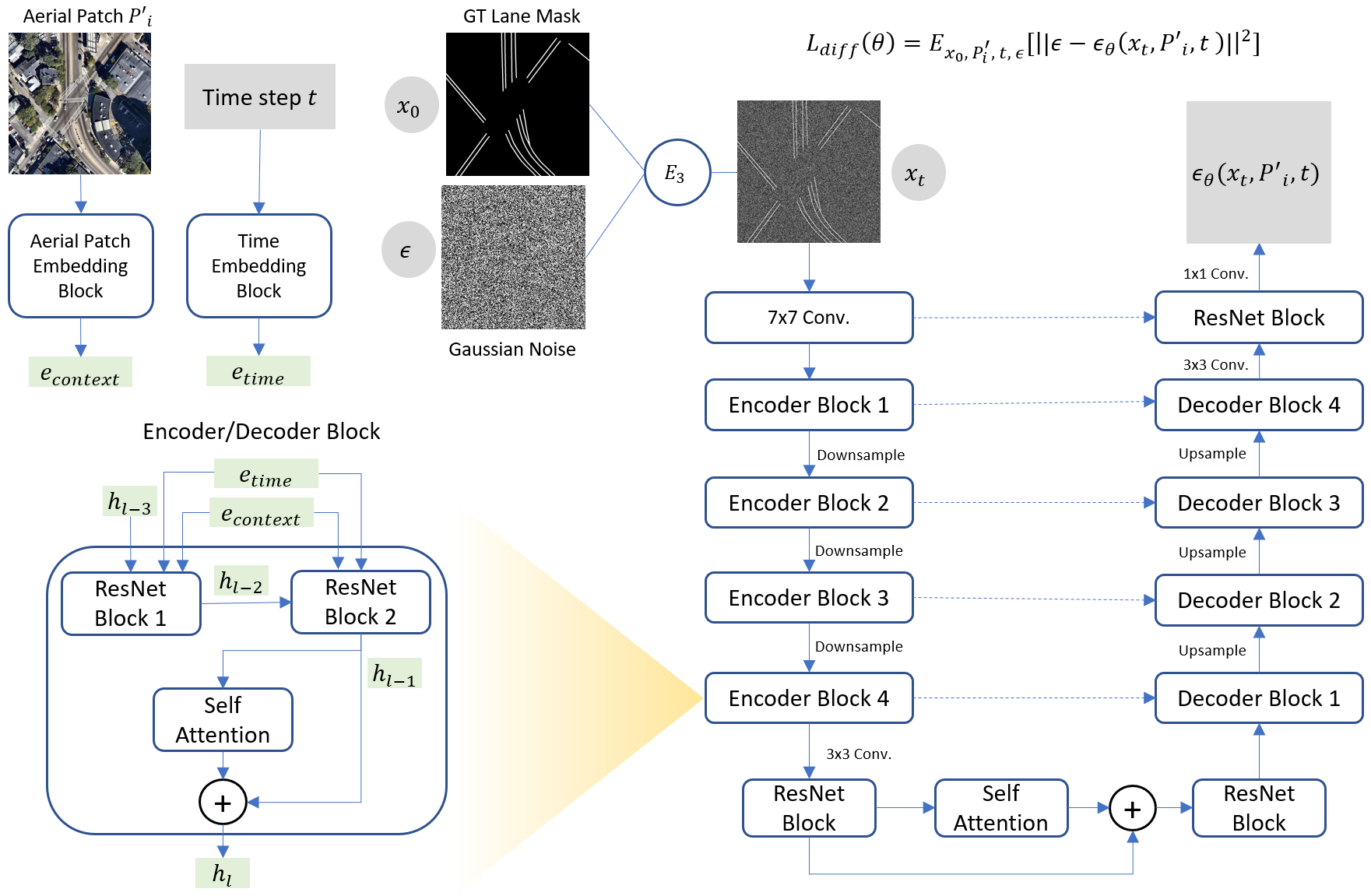}%
}
%
%

\subfloat[ ]{%
  \includegraphics[clip,width=.95\textwidth]{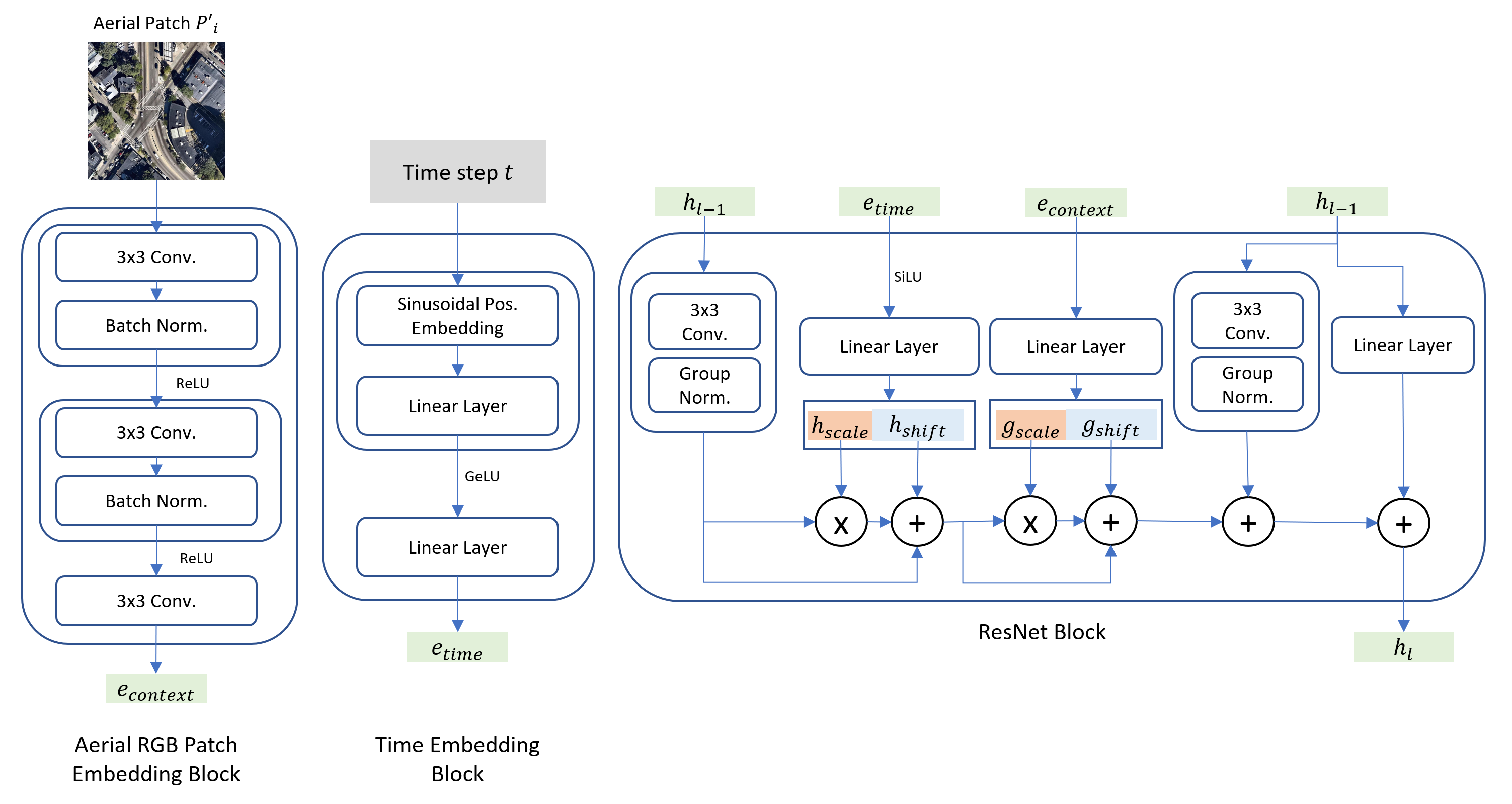}%
}
\caption{\textls[-5]{Overall architecture based on the framework from Improved DDPM~\cite{nichol2021improved}. Noise is progressively added to the segmentation mask through the forward process, as described in Equation~\eqref{eq:x-t-equation}. The U-Net is then trained to learn the denoising process using the loss function in Equation~\eqref{eq:diffusion-loss}. The aerial patch and time-step are processed through embedding blocks, with the resulting embeddings injected into all ResNet~\cite{he2016deep} blocks except the final one, where only the time embeddings are used. (\textbf{a)} 
 U-Net~\cite{ronneberger2015u} architecture for learning the denoising process. Left side: inner structure of encoder--decoder block. Right side: U-net architecture with stacks of encoder--decoder blocks. (\textbf{b}) Inner structure of the Aerial Patch and Time Embedding blocks, and the ResNet block from Figure~\ref{fig:4}a. } \label{fig:4}}
\end{figure}

Initially, in the Second Stage, we preprocess the training dataset similarly to the previous stage, applying cropping, random rotations, and random adjustments to color and brightness, followed by rescaling. As in the previous stage, these data augmentation techniques also enhance the generalization capabilities of the model. We let \(\mathcal{R} = \{ (s'_i, P'_i)\}_{i=1}^{n}\) represent the preprocessed dataset, where \(s'_i\) is the ground truth (GT) segmentation mask and \(P'_i\) is an aerial patch, initially sized \(1024 \times 1024\) pixels (as in the First Stage), subsequently rescaled to \(256 \times 256\) to fit GPU memory constraints. Then, we encode the rescaled aerial patches \(P'_i\) using a simple three-layer CNN and generate timestep embeddings through a block composed of sinusoidal positional embeddings~\cite{vaswani2017attention}, widely used in diffusion models, combined with two linear layers (see embedding blocks in Figure~\ref{fig:4}). We then utilize the rescaled GT segmentation masks  to conduct the forward diffusion process of a DDPM~\cite{ho2020denoising}, producing noisy segmentation masks guided by a sigmoid variance schedule~\cite{jabri2022scalable}, used for training stability. Finally, we train a U-Net~\cite{ronneberger2015u} to learn the denoising process, following the parametrization introduced by DDPM~\cite{ho2020denoising}. The U-Net architecture is widely utilized in diffusion models for image processing due to its skip connections between the encoder and decoder, which effectively capture global context and restore fine details from noisy inputs. The aerial RGB patches and timestep embeddings are injected into the U-Net via ResNet blocks~\cite{he2016deep}. The overall architecture for learning the denoising process of the diffusion model closely follows the design principles of Improved DDPM~\cite{nichol2021improved}. The U-Net is trained with the following loss:

\begin{equation}
  \mathcal{L}_{diff} (\theta) = \mathbb{E}_{x_0, P'_i, t, \epsilon} [\ ||\epsilon - \epsilon_{\theta} (x_t, P'_i, t)||^2 ]\ ,
  \label{eq:diffusion-loss}
\end{equation}
where \(x_0\ \sim q(x_0) \), \(q(x_0)\) is the GT segmentation masks data distribution,  and \(P'_i\) is the corresponding rescaled aerial patch of the sample \(x_0\), with \( x_0, P'_i \in \mathbb{R}^{256 \times 256} \), \(t \sim U([1, T])\)  is the time-step sampled from the uniform distribution \( U \), \( T \) is the length of horizon for the forward diffusion process,  \(\epsilon \sim \mathcal{N}(\boldsymbol{0}, \boldsymbol{I}) \)  and \( x_t \) is computed from \( x_0 \)  by means of the following equation (as in DDPM): 
\begin{equation}
    x_t = \sqrt{\bar{\alpha}_t} x_0 + \sqrt{1-\bar{\alpha}_t} \epsilon,
    \label{eq:x-t-equation}
\end{equation}
where \(\bar{\alpha}_t := \prod_{s=1}^t\alpha_s \), with \( \alpha_t := 1-\beta_t \) and \( \beta_t \)s are taken from the sigmoid variance schedule~\cite{jabri2022scalable}.

Once the diffusion model is trained, we follow the DDIM sampling procedure introduced by Song et al.~\cite{song2020denoising}, where samples can be generated through a deterministic process from the initial point \(x_T\) to the final point \(x_0\), i.e., no noise is added during intermediate steps. Since DDIM employs a non-Markovian process, sampling is significantly faster compared to standard DDPM sampling. Although this typically results in lower sample quality, our case is less affected because we do not start from Gaussian noise. Additionally, in DDIM, the number of training forward steps is independent of the number of sampling steps, which allows sampling from only a selected subset during the inference phase. We choose a subset \(\tau =\{x_{\tau_1}, x_{\tau_2}, \ldots, x_{\tau_S}\}\), where \(\{\tau_1, \tau_2, \ldots,  \tau_S\}\) is an increasing subsequence of \(\{1, 2, \ldots, T\}\) of length \(S\), such that the difference between \(x_{\tau_{j}}\) and \(x_{\tau_{j-1}}\) is constant, with \(j \in \{2,3, \ldots , S\}\). This constant step size makes the sampling process more stable by avoiding large jumps between steps while still maintaining efficiency. The resulting sampling trajectory is \(\{x_{T}, x_{T-\delta}, x_{T-2\delta}, \ldots, x_{T-(S-1)\delta}, x_0\}\) (the reverse of \(\tau\)), where \(\delta\) is the step size, \(S\) is the total number of sampling steps, and \(S\delta = T\). Each pair \(x_{t}\) and \(x_{t-\delta}\) is connected through the following equation, which corresponds to a special case of Equation (12) in DDIM~\cite{song2020denoising}, with \(\sigma_t = 0\):

\begin{equation}
    x_{t-\delta} = \sqrt{\bar{\alpha}_{t-\delta}} \left( \frac{x_t - \sqrt{1-\bar{\alpha}_t} \ \epsilon_{\theta}^t}{\sqrt{\bar{\alpha}_t}} \right) + \sqrt{1-\bar{\alpha}_{t-\delta}} \ \epsilon_{\theta}^t,
    \label{eq:x-t-delta-equation}
\end{equation}
where \(\epsilon_{\theta}^t\) is the output of U-Net at time-step \(t\) and the rest of terms are defined above. In DDIM, \(\sigma_t \) controls the amount of noise added during sampling, and when it is set to 0, the process is no longer random but deterministic.

To cover the entire testing tiles, we adopt the same sliding window approach used in the First Stage, employing a fixed patch size of \( 1024 \times 1024\) with identical vertical and horizontal strides of 512 to refine each sliding window. However, at this stage, the diffusion model receives two inputs: the aerial patch and its corresponding unrefined segmentation mask, both resized to \( 256 \times 256\) to match the input size of the model. The refined mask output is then resized back to the original patch dimensions of \( 1024 \times 1024\). Once the entire tile is processed, we compute the average of the overlapping regions to obtain the final refined segmentation mask (see Figure~\ref{fig:sliding-window-prediction}). As in the previous stage, averaging is used to enhance robustness, since overlapping patches may include different contextual information.

\subsection{Third Stage: Segmentation-to-Graph Algorithm}

In the Third Stage, we follow the approach used in LaneExtraction~\cite{he2022lane}. Specifically, we apply a threshold \(\alpha = 0.5\) to the refined segmentation masks (interpreted as probability maps) to obtain binary masks. These masks are then processed using a morphological thinning algorithm~\cite{guo1989parallel} to produce a skeleton structure, which is subsequently converted into a graph. Morphological thinning simplifies the mask to a skeleton, preserving the essential structural features of the lanes while eliminating redundant pixels. A post-processing step prunes the graph by removing small connected components and spurs (defined as edges extending from a node without connecting back to the main graph structure)
. Finally, the Douglas–Peucker algorithm~\cite{ramer1972iterative} is employed to simplify the graph, reducing the number of nodes for improved clarity and computational efficiency while preserving its overall shape.

\subsection{Implementation Details}
\label{sec:experimental-details}

The architecture for the D-LinkNet~\cite{zhou2018d} in the First Stage was implemented in TensorFlow 2~\cite{tensorflow2015-whitepaper}. We used the same network hyperparameters as LaneExtraction~\cite{he2022lane}. To be more precise, we trained the model for 500 epochs, starting with a learning rate of \(1 \times 10^{-3}\) 
 and reducing it by a factor of 10 at epochs 350 and 450 and employed the AdamW~\cite{loshchilov2017decoupled} optimizer. The architecture of the D-LinkNet was adopted from prior work (refer to Figure 3 in LaneExtraction~\cite{he2022lane}). Data augmentation techniques, including random cropping, rotations, and random modifications to color and brightness, were employed. We used patch sizes of \(1024\times1024\) for the pipeline of our method. The batch size was set to 8.

The architecture for the diffusion model in the Second Stage was implemented in Pytorch 2~\cite{paszke2019pytorch}. We used  the sigmoid variance schedule~\cite{jabri2022scalable} with hyperparameters set to \(start = -3\), \(end = 3\), and \(\tau = 1\). The length of the forward process was set to \(T = 1000\).  The model was trained for \(200K\) steps, utilizing a learning rate of \(8 \times 10^{-5}\) with the Adam optimizer~\cite{kingma2014adam} and applying an exponential moving average (EMA) with decay factor of 0.995. The dimensions for both time and aerial patch embeddings were set to 256. Segmentation masks utilized one channel, while aerial RGB patches employed three channels. The batch size was set to 8. Data augmentation strategies included random cropping, horizontal and vertical flipping, integer rotations within \([-15, 15]\) degrees, and color jitter for the aerial RGB patches, which were sized to \(1024\times1024\). As outlined in Section~\ref{sec:segmentation-refinement-stage}, the patches were resized to \(256\times256\) to fit into the GPU memory. Our implementation for the diffusion model is based on the following git repository: \url{https://github.com/lucidrains/denoising-diffusion-pytorch} (accessed on 30 July 2025 
). This implementation uses the v-parametrization (as explained in \mbox{Appendix D} in~\cite{salimans2022progressive}) to train the diffusion model; we left it as it is. 

The training and inference of both models were carried out on a single NVIDIA GeForce RTX 3090 GPU using CUDA Version 12.2.

\section{{Results}}
\label{sec:results}

\subsection{Dataset}
\label{sec:dataset}
We use the dataset introduced by LaneExtraction~\cite{he2022lane}. The dataset consists of 35 aerial tiles of size \( 4096 \times 4096 \) covering the area of four cities in United States:  Miami, Boston, Seattle, and Phoenix. The ground sample distance (GSD) of the aerial images is 12.5 cm per pixel. {To ensure a fair and consistent comparison with prior work~\cite{he2022lane}, we adopt the same dataset split they used: 24 tiles for training and 11 for testing, corresponding to approximately 70\% and 30\% of the data, respectively. This ratio is widely adopted in deep learning-based computer vision tasks as it offers a practical balance between providing sufficient data for learning and maintaining a robust test set for evaluation. Additionally, by ensuring that all four cities (Miami, Boston, Seattle, and Phoenix) are represented in both subsets, we preserve geographic diversity during training and avoid biasing the model toward specific urban areas. This split supports assessing the model’s ability to generalize across varied urban environments while aligning with established benchmarks.}

For each tile, the ground truth lane graph is known. The corresponding ground truth segmentation masks are generated by rendering 5-pixel-wide white lines on black backgrounds along the directed edges of the ground truth graph. This width is chosen to provide sufficient contextual information while avoiding overlaps with adjacent lanes. The ground truth direction maps are created in a similar fashion, but colored lines in BGR (Blue--Green--Red) format are used to encode the direction of the edges, the first and second components of the color encode the \( x\) and  \( y\) normalized directions, respectively, whereas the third component is kept constant, pixels without lanes are set to the zero vector. Then, a training tile triple is formed, consisting of the aerial tile, the ground truth segmentation mask, and the ground truth direction map. After the training tile triples are formed, we create the training set that is used for the CNN (First Stage) and the diffusion model (Second Stage) by using a sliding window of size  \( 2048 \times 2048 \) with vertical and horizontal stride of 1024 on each training tile. This results in 216 training samples. It is important to note that the sliding window size used to create the training tiles differs from the patch size used to generate the training patches input to the model. The training patches, sized \(1024 \times 1024\), are random crops extracted from training tiles of size \(2048 \times 2048\). The chosen stride used to create the training tiles offers a good balance between generating a sufficient number of training samples and avoiding overly similar examples. As mentioned in Section~\ref{sec:materials-and-method}, the testing tiles are not cropped for inference but used in full size (\(4096 \times 4096\)). A sliding window of the same size as the training patches (\(1024 \times 1024\)), with horizontal and vertical strides of \(512\), is used to cover the entire tile. This methodology, with minor changes, 
is also employed in LaneExtraction~\cite{he2022lane}.

\subsection{Evaluation Metrics}
\label{sec:metrics}

We employ the GEO and TOPO metrics to evaluate our method. These metrics are adapted from the problem of road extraction~\cite{biagioni2012inferring}. To ensure a fair comparison, we adopt the methodology and hyperparameters (used by the metrics) described in LaneExtraction~\cite{he2022lane} across all experiments.

\subsubsection{GEO Metrics}

This methodology consists in densifying the predicted lane graph \(\hat{G}_0 = \{\hat{V}_0, \hat{E}_0\}\) and the ground truth lane graph \(G_0 = \{V_0,E_0\}\), where \(\hat{V}_0\) and \( V_0\) are sets of vertices and \( \hat{E}_0 \) and \(E_0\) are sets of edges of the respective graphs. The graphs are densified by inserting additional nodes such that the distance between any two connected nodes is less or equal than 25 cm in real-world distance (2 pixels). For instance, if two nodes are 100 cm apart, two intermediate nodes would be added. This interpolation makes the evaluation metrics more robust, especially in cases where the original graphs contain few nodes. The densification process induces a  predicted lane graph \(\hat{G} = \{\hat{V}, \hat{E}\}\) from \(\hat{G}_0\) and a ground truth lane graph \(G = \{V,E\}\) from \(G_0 \), where \(\hat{V}\) and \( V\) are sets of vertices and \( \hat{E} \) and \(E\) are sets of edges of the respective graphs. The pair \( (\hat{v}, v) \), with \( \hat{v} \in \hat{V} \) and \( v \in V \), is considered a valid match if \( ||\hat{v} - v||_2 < r \) (as in Section~\ref{sec:materials-and-method}, presence or absence of a caret \(\hat{\enspace}\) on a symbol indicates prediction or ground truth values, respectively), where \( r \) determines the tolerance of the metric. We set \( r = 1 m \) in real-world distance (8 pixels). This allows the metric to have a reasonable failure tolerance. After that, a maximal one-to-one match between \(\hat{V}\) and \(V \) is computed and the matched vertices in the predicted lane graph are denoted as \(\hat{V}_{match}\). Then, the three GEO metrics, precision, recall and F1 score, are defined as follows:

\begin{equation}
  pre_{GEO} = \frac{|\hat{V}_{match}|}{|\hat{V}|}, \ \ rec_{GEO} = \frac{|\hat{V}_{match}|}{|V|}, \ \ F_{1GEO} = 2 \cdot \frac{ pre_{GEO} \cdot rec_{GEO} }{pre_{GEO} + rec_{GEO}}.
  \label{eq:geo-metrics}
\end{equation}

\subsubsection{TOPO Metrics}
GEO metrics do not take into account connectivity, so TOPO metrics aim to overcome this limitation. TOPO metrics are built on top of GEO metrics. For each matched vertex pair \( (\hat{v}, v) \) in the GEO metrics, the subgraphs \(S_{\hat{v}}\) and \(S_v\) are defined in the following way: \( S_{\hat{v}} = \{\hat{u} \in \hat{G}:  d(\hat{v},\hat{u}) < 50m \} \) and \( S_v = \{u \in G: d(v,u) < 50m \} \), where \( d(a,b)\) is the length (accumulated sum of Euclidean distances between nodes along the path from \(a\) to \(b\)) of the shortest path from node \(a\) to node \(b\), if there is any, otherwise it is \(\infty\).
Then, the GEO metrics between the two subgraphs \(S_{\hat{v}}\) and \(S_v\), denoted as \(pre_{GEO}( S_{\hat{v}}, S_v)\) and  \(rec_{GEO}(S_{\hat{v}}, S_v)\), are computed for each vertex pair \( (\hat{v}, v) \), and the final TOPO metrics are defined as follows:
\begin{equation}
pre_{TOPO} = \frac{\sum_{(\hat{v}, v)} pre_{GEO}(S_{\hat{v}}, S_v)}{|\hat{V}|},
\label{eq:topo-precision}
\end{equation}

\begin{equation}
rec_{TOPO} =  \frac{\sum_{(\hat{v}, v)} rec_{GEO}( S_{\hat{v}}, S_v)}{|V|},
\label{eq:topo-recall}
\end{equation}

\begin{equation}
F_{1TOPO} = 2 \cdot \frac{ pre_{TOPO} \cdot rec_{TOPO} }{pre_{TOPO} + rec_{TOPO}}.
\label{eq:topo-recall}
\end{equation}

\subsubsection{{Strengths and Limitations of Evaluation Metrics}}
{These metrics are robust to small variations between the ground truth and predicted nodes as they do not require an exact geo-positional match. Instead, they consider a match when nodes fall within a specified radius, with the tolerance hyperparameter controlling this threshold. When used together, they offer a comprehensive performance analysis: precision highlights the accuracy of true positive predictions, recall measures the model’s ability to minimize false negatives, and the F1 score balances both metrics, providing a holistic view of the model performance. However, they have several drawbacks as well. First, post-processing is required, including graph densification and computing the optimal one-to-one match between the predicted and ground truth graphs. Additionally, GEO metrics do not account for graph connectivity, since they focus solely on matching nodes at the geo-positional level while disregarding whether those nodes are actually connected, a feature of the lane graph that is highly relevant for downstream applications. While TOPO metrics attempt to address this limitation, they can be computationally expensive, particularly when dealing with large graphs, since they require computing GEO metrics over many subgraphs. Another challenge is selecting appropriate hyperparameters, which can be difficult, particularly when dealing with diverse geographical areas and multiple image resolutions.}

\subsection{Baselines}
\label{sec:baselines}

In this section, we describe the baselines used to compare our method. We evaluate our approach against three methods: a standard U-Net~\cite{ronneberger2015u}, LaneExtraction~\cite{he2022lane}, and MedSegDiff~\cite{wu2022medsegdiff} (an ensemble of diffusion models). U-Net is a widely used network for image segmentation across various domains. LaneExtraction is specifically designed to extract lane segmentation masks from aerial imagery, utilizing a D-LinkNet~\cite{zhou2018d} with two heads to predict both lane segmentation masks and direction maps. MedSegDiff was originally designed for medical image segmentation using diffusion models conditioned on the input images, with the corresponding segmentation masks as the targets for diffusion. In our case, however, we adapt this approach by using lane segmentation masks and aerial images as inputs instead of medical imagery. To enhance robustness, the sampling procedure is repeated multiple times, effectively forming an ensemble of diffusion models. We select two methods based solely on CNNs and one method based solely on diffusion models to demonstrate that our hybrid approach, which combines both, is more effective than either method individually. Another noteworthy aspect is that CNN-based methods are deterministic, producing consistent results across multiple runs, whereas methods involving diffusion models may yield slightly different outputs each time, requiring multiple runs for a more reliable evaluation.

After obtaining the lane segmentation masks, all methods, including ours, apply the same segmentation-to-graph algorithm to obtain the final lane graph. As mentioned earlier, the quality of the final lane graph is directly influenced by the accuracy of the lane segmentation masks. Since our focus is on the lane graph, we compute the metrics based on it. However, because the segmentation-to-graph algorithm is identical across all methods, this evaluation serves as an indirect measure of the quality of the lane segmentation masks produced by each method.

\subsection{Quantitative Results}
\label{sec:quantitative-results}

In this section, we report the experimental results obtained on the publicly available dataset introduced in Section~\ref{sec:dataset}, comparing our method against the three baseline approaches. Additionally, we include ablation studies to analyze the contribution of each component in our pipeline. Performance is assessed using the GEO and TOPO metrics, focusing specifically on undirected lane graphs in non-intersection regions. As discussed in Section~\ref{sec:introduction}, the extraction and evaluation of lane graphs in intersection areas remain an open challenge and are deferred to future work.

Table~\ref{tab:graph-metrics} shows the comparison of our method against the baselines. Our method outperforms all baselines in both GEO and TOPO \(F_1\) scores. Regarding methods purely based on CNNs, our method improves the GEO and TOPO \(F_1\) scores by approximately \(1.5\%\) and \(3.5\%\), respectively, compared to LaneExtraction~\cite{he2016deep}, the second best performing method. Compared to the ensemble of diffusion models~\cite{wu2022medsegdiff}, our method achieves a \(28\%\) gain in the GEO F1 score and a \(34\%\) improvement in the TOPO F1 score. These two metrics are the primary indicators for assessing lane graph quality. The most notable improvement occurs in the TOPO \(F_1\) score (\(3.5\%\)), which evaluates lane graph connectivity. Our conditional diffusion model effectively reduces false negatives without introducing false positives. This is reflected in a significant increase in recall (\(9\%\) improvement with respect to LaneExtraction) while maintaining comparable precision levels after applying the diffusion model, leading to overall improved \(F_1\) scores.

In our ablation studies, we set \(S=25\) DDIM sampling steps as the default setting. This value has been used in previous studies employing diffusion models for image segmentation~\cite{wu2022medsegdiff, wu2023medsegdiff}. Each diffusion model experiment was repeated 10 times for statistical reliability, with the corresponding means and standard deviations reported in Tables~\ref{tab:components-ablation}--\ref{tab:noise-levels-ablation}. 

Table~\ref{tab:components-ablation} presents an ablation study of the key components of our method: (1) the conditional DDPM~\cite{ho2020denoising}, where the training of the diffusion model is conditioned on RGB aerial patches, and (2) the conditional DDIM~\cite{song2020denoising}, where the initial latent variable \(x_T\), for the sampling procedure, is conditioned on the unrefined segmentation mask produced by the CNN. As discussed earlier, we evaluate three types of conditioning strategies (refer to Section~\ref{sec:segmentation-refinement-stage}) for the DDIM sampling; however, we consider the default setting to be the addition of Gaussian noise \(\epsilon \sim \mathcal{N}(\boldsymbol{0}, \boldsymbol{I})\) to the unrefined segmentation mask. When the conditional DDPM is not used, the diffusion model is trained only on the segmentation masks without any conditioning. Furthermore, when the conditional DDIM is not used, the initial latent variable is assumed to be sampled from a standard normal distribution, i.e., \(x_T \sim \mathcal{N}(\boldsymbol{0}, \boldsymbol{I})\). It is important to note that the conditional DDPM component is used during training, while the conditional DDIM is used for sampling. {In the table, we observe that a diffusion model trained solely on lane segmentation masks (first row), without any conditioning, performs quite poorly across all metrics (nearly zero). This outcome is expected, since the generation is not conditioned on anything, and although the generated masks may resemble lanes, they are scattered arbitrarily, lacking any alignment with the corresponding aerial images. From the two key components of our model, the conditional DDIM has the greater impact on the overall performance. Notably, even without using the conditional DDPM during training, the conditional DDIM alone can still yield reasonable results. This can be explained by the fact that the lane masks produced by the CNN are quite a good approximation, and this imposes a strong prior for the inference of the diffusion model. On the other hand, the conditional DDPM also provides a prior during training, which helps to refine the lane masks produced by the CNN during inference. As a result, the best performance is achieved when both components are used together. This highlights the importance of both conditioning mechanisms and supports our design choice of incorporating both into the diffusion model.}

Table~\ref{tab:noise-ablation} outlines an ablation study on the effect of adding Gaussian noise (\(\epsilon \sim \mathcal{N}(\boldsymbol{0}, \boldsymbol{I})\)), to the unrefined segmentation mask while varying the number of DDIM sampling steps. In the table, we observe marginal gains in the GEO \(F_1\) score (\(+0.002\) on average) compared to starting the DDIM sampling directly from the unrefined segmentation mask. Conversely, the TOPO \(F_1\) score does not consistently improve with the addition of Gaussian noise, though the declines when it performs worse are minimal, \(-0.002\) when \(S=50\) and \(-0.001\) when \(S=250\). Although the improvements in \(F_1\) scores are marginal, adding Gaussian noise reduces the gap between precision and recall compared to not adding it, suggesting that the model achieves a better balance between avoiding false positives and false negatives. We hypothesize that adding Gaussian noise to the unrefined segmentation mask produces an initial latent variable \(x_T\) that more closely resembles a sample from a Gaussian distribution (the type of variable expected during DDIM sampling), which may help stabilize the reverse diffusion trajectory compared to using the unrefined mask directly as the initial variable.

In Table~\ref{tab:noise-levels-ablation}, we present an ablation study on the impact of adding different noise levels to the unrefined segmentation mask using forward diffusion steps (Equation~\eqref{eq:x-t-equation}). Our goal is to analyze how the model performs during sampling when applying the same noise injection strategy used during training, but using the unrefined segmentation mask instead of the ground truth as the clean data sample. We control the noise level by varying the number of forward steps (FS) from 0 (0\%) to 1000 (100\%), as in the training phase, while keeping the number of sampling steps fixed at \(S=25\). The best results are achieved when 50\% noise is applied to the unrefined mask, with only minor performance variation observed up to 75\%. Notably, model performance drops by more than \(7\%\) in the GEO \(F_1\) score and \(11\%\) in the TOPO \(F_1\) score at 90\% noise. At 100\% noise (i.e., \(x_T\) consisting of pure Gaussian noise), there is a pronounced decline, with over \(20\%\) and \(27\%\) drops in GEO and TOPO \(F_1\) scores, respectively. These results highlight the limitation of the model in reconstructing the segmentation mask when starting from pure Gaussian noise (see the underlined results in Table~\ref{tab:noise-levels-ablation}).

In Table~\ref{tab:computational-resources}, we present an analysis of the computational resources used by our method and the baselines. While our method requires approximately 11 times more training time than the CNN-based approaches, it only demands about three times as much inference time and memory compared to LaneExtraction, the second best performing method. This is expected, as our method incorporates a CNN component along with a diffusion model, which typically require substantial training to converge. However, this overhead could be mitigated through parallel training on multiple GPUs, which significantly reduces the training time. Compared to the other diffusion-based model, our method requires twice as much training time and approximately 1.5 times more memory. However, it achieves significantly faster inference, about 18 times faster than the ensemble of diffusion models (DM)~\cite{wu2022medsegdiff}. It is also worth noting that the inference time of our method can be further reduced by decreasing the number of inference steps. For example, reducing the default from 25 to 10 steps still yields strong performance, as demonstrated in Table~\ref{tab:noise-ablation}.

Additionally, we carried out some experiments to analyze pixel-level changes resulting from refining the lane masks with the diffusion model (Second Stage of our method), as presented in Appendix~\ref{app:pixel-statistics}.

\subsection{Qualitative Results}
\label{sec:qualitative-results}

Figure
~\ref{fig:complex-cases} illustrates several challenging scenarios in which CNNs fail to accurately segment lanes from aerial imagery. These complex cases include occlusion caused by queues of cars (first row) or trees covering the road (second row), changes in road texture (third row), and lighting variation (fourth row). As observed in the figure, CNNs produce incomplete masks, as they struggle to detect lanes when the visual pattern does not resemble a road or when the road is barely visible. In contrast, our diffusion model successfully produces sharp and complete lane masks by refining the initial masks predicted by the CNN, as shown in the third column.

Figures
~\ref{fig:results-tile-A} and~\ref{fig:results-tile-B} present a comparative visualization of three methods: LaneExtraction~\cite{he2022lane} (top row), an ensemble of diffusion models~\cite{wu2022medsegdiff} (middle row), and our method (bottom row), applied to two regions from different evaluation tiles. While LaneExtraction struggles with fragmented lane segments (red dotted box, first row) due to challenges inherent in aerial imagery (refer to Figure
~\ref{fig:complex-cases}), the diffusion ensemble produces blurry masks (red dotted box, second row), i.e., the segmented pixels do not form continuous lanes with well-delineated boundaries, as lateral pixel variations across the diffusion ensemble blur the averaged output. By contrast, our method fills the gaps in the fragmented lanes left by LaneExtraction and produces sharp masks, i.e., masks with clearly defined lane boundaries, by training a conditional DDPM conditioned on aerial RGB patches and conditioning the initial latent variable for DDIM sampling on the output of a CNN, as detailed in Section~\ref{sec:segmentation-refinement-stage}. These visual results show a significant reduction in false positives and false negatives in the resulting lane graphs (third column), particularly excelling at removing small, incorrect segments of nodes. 

\section{Discussion}
\label{sec:discussion}
Our experiments demonstrate that integrating a CNN with a diffusion model, where the CNN's lane segmentation prediction conditions the initial latent variable in the sampling procedure of a conditional diffusion model, yields complete and sharp lane masks, surpassing the capabilities of either model in isolation (see Figures~\ref{fig:results-tile-A} and~\ref{fig:results-tile-B}). This enhancement leads to higher-quality lane graphs extracted from the lane masks, with a notable improvement in connectivity, as reflected by the TOPO \(F_1\) score  (refer to Table~\ref{tab:graph-metrics}). Furthermore, by analyzing the impact of varying noise levels applied to the initial unrefined segmentation mask during DDIM sampling, we found that while introducing Gaussian noise enhances the stability and robustness of the method, it does not lead to significant improvements in quantitative metrics.

The lane graph extracted from the refined lane masks output by the diffusion model can be viewed as a denoised version of the lane graph produced by the CNN, enhancing the overall quality of the extracted lane graph.  However, several limitations are also apparent in the results: (1) our method struggles to replace large false negative segments (e.g., the prominent red segment at the bottom right bifurcation in Figure~\ref{fig:results-tile-A}), and (2) it sometimes generates node segments with significant positional shifts, causing mismatches (visible in Figure~\ref{fig:results-tile-A} as overlapping blue--red segments near the same bifurcation). Future research could address these two key challenges: (1) restoring complete segments when the CNN provides no cues for the diffusion model and (2) reducing positional shifts to prevent mismatches in the final lane graph. {Potential solutions to Challenge (1) include designing specialized loss functions for the CNN that focus on large occluded regions, as well as increasing the number of training samples that represent such scenarios. Additionally, incorporating contextual cues from nearby lanes could help the diffusion model infer missing segments by leveraging the structure of the surrounding road network. Introducing auxiliary supervision—such as occlusion masks or semantic segmentation maps (e.g., of trees or bridges)—could further assist the CNN in handling these areas. Pretraining or multitask learning on related tasks, such as road topology estimation, could also enhance the CNN's ability to generalize in the presence of large, continuous occlusions. Finally, applying a graph completion step during post-processing may help recover missing connections by exploiting structural patterns in the predicted lane graph. To address Challenge (2), one potential approach is to widen the ground truth lane segmentation masks used by the CNN or to incorporate additional geometric information—such as distances or other geospatial properties between lanes—as input to both models. An alternative strategy could involve introducing a third model that refines the predicted lane graph using geometric cues, such as lane curvature, or applying geometric losses to regress the nodes toward their correct positions.}

While our method currently focuses on undirected lane graphs in non-intersection areas, we believe that extending to directed lane graphs across all road areas represents a promising research direction as well. {To estimate the direction of lanes, we can utilize the direction map predicted by the CNN (first step) and determine the most likely direction using a heuristic based on edge orientations within a lane segment (similar to Equation~(2) in~\cite{he2022lane}). To extend our method to intersection areas, we consider several possible strategies. We first recall that terminal nodes are defined as those with exactly one incoming or outgoing edge within lanes in non-intersection areas. We then define intersection paths as sets of connected nodes that link two lanes from non-intersection areas.
The following list outlines three such approaches:
\begin{itemize}
    \item Iterative Key Node Prediction: 
    One approach is to use a model that predicts key nodes within intersection areas, such as nodes derived from Bézier curves connecting terminal nodes. An iterative algorithm can then use visual context and the current node position to predict the next node, starting from a terminal node and stopping when reaching another terminal node.
    \item Joint Segmentation and Diffusion Model: Another alternative is to predict segmentation masks of intersection paths and merge them with the lane segmentation masks from non-intersection regions. Then, we could apply a conditional DDIM that takes both segmentation masks into account, allowing the model to connect all lane segments in a unified manner.
    \item Terminal Node Pairing: A third approach involves training a model to identify pairs of terminal nodes that should be connected within an intersection. Intersection areas could be detected using additional segmentation masks. Once a pair is identified, another model could predict the intermediate nodes and edges that connect the two terminal nodes. This approach is similar to the one proposed in LaneExtraction~\cite{he2022lane}.
\end{itemize}}

It is also worth noting that our method can be extended to other domains, such as biomedical imaging for vessel segmentation, contour line extraction, or any application requiring sharp and complete segmentation of thin structures. Potential extensions could include integration with stroke-based rendering techniques~\cite{nolte2022stroke} or value-function-guided segmentation~\cite{melnik2021critic}.

\section{Conclusions}

In this paper, we introduce a novel approach for extracting sharp and complete lane segmentation masks from aerial imagery. Our method leverages conditional diffusion models along with a novel conditioning strategy for the initial latent variable used in the sampling procedure of the diffusion model. A CNN is used to produce an initial approximation of the lane masks, which are subsequently refined through the diffusion sampling procedure. These refined masks are then processed by a conventional rule-based segmentation-to-graph algorithm to construct lane graphs in non-intersection areas of aerial imagery. As shown by our quantitative and qualitative results, this refinement process significantly enhances the quality of the final lane graphs, particularly in terms of connectivity, a critical factor for downstream applications such as autonomous driving.

\vspace{6pt}




\authorcontributions{A.R. drafted and polished the manuscript, analyzed the data, developed the methodology, implemented the code and performed the experiments. A.M. developed the methodology, provided technical guidance, analyzed the results and revised the manuscript. N.S. helped in the design of the methodology, analyzed the results and revised the manuscript. D.W. and Y.Z. 
 provided technical guidance, managed the project and revised the manuscript. H.R. supervised the project and revised the manuscript. All authors have read and agreed to the published version of the manuscript.}

\funding{This research received no external funding.}

\institutionalreview{Not applicable.}

\informedconsent{Not applicable.}

\dataavailability{The dataset for all the experiments is available in the following link: \url{https://github.com/songtaohe/LaneExtraction/tree/master/dataset} (accessed on 30 July 2025
).}

\acknowledgments{During 
 the preparation of this manuscript, the authors used DeepSeek-V3 and ChatGPT-4o for the purposes of grammar checking and text refinement. The authors have reviewed and edited the output and take full responsibility for the content of this publication.}

\conflictsofinterest{The authors 
 declare no conflicts of interest.} 



\appendixtitles{yes} 
\appendixstart
\appendix

\section[\appendixname~\thesection]{Pixel Statistics}

\label{app:pixel-statistics}

In 
Table~\ref{tab:stats-absolute-pixel-changes}, we detail the number of pixels changed from white to black and vice versa, varying the number of sampling steps and comparing scenarios with and without noise. This table demonstrates the capability of the diffusion model to perform changes in both directions, highlighting its utility not just in filling gaps between lane segments but also in rectifying misplaced white pixels in the unrefined segmentation mask. The third column, which indicates absolute difference between the changes from white to black pixels and vice versa, reveals that the model maintains a relative balance in both scenarios, where the differences represent about one fifth of the total in the first column and approximately one fourth in the second column. Additionally, Table~\ref{tab:stats-percentage-pixel-changes} displays the percentages of relative pixel changes from white to black and vice versa, using the same settings as Table~\ref{tab:stats-absolute-pixel-changes}. This table reinforces the idea illustrated in Table~\ref{tab:stats-absolute-pixel-changes} that the diffusion model changes pixels in both directions. The percentages of relative changes from white to black are significantly higher than their counterparts, reflecting the overall scarcity of white pixels, which are restricted to the 5-pixel-wide lines representing the lanes.
 
\begin{table}[H]
\caption{Average number of pixels changed from the unrefined to the refined segmentation masks on the test dataset. \(S\) represents the number of DDIM sampling steps. The third column shows the absolute difference between the first and second columns. The format of the results indicates the mean \(\pm\) std. across 11 testing tiles in one repetition of each experiment. The testing tiles contain \(4096\times4096\) pixels.}
\label{tab:stats-absolute-pixel-changes}

\begin{tabularx}{\textwidth}{cCCCC}
\toprule
\multirow{2}{*}{\boldmath \(S\) } & \multirow{2}{*}{ \textbf{Noise} } & \multicolumn{2}{c}{\ \textbf{Mean of \# of Changed Pixels} \boldmath\(\times 1K\) \ } & \multirow{2}{*}{\textbf{Abs. Diff. (}\boldmath\(\times 1K\)\textbf{)}} \\ \cmidrule{3-4}
                         &                         & \textbf{White}\boldmath \(\to\) \textbf{Black} & \textbf{Black}\boldmath \(\to\) \textbf{White} &  \\ \midrule
\multirow{2}{*}{10} & \checkmark & \(128.106 \pm 14.974\)  & \(103.657 \pm 25.401\) & \(28.322 \pm 13.898\) \\ 
                    & \xmark & \(121.853 \pm 13.809\)  & \(102.733 \pm 27.849\) & \(27.377 \pm 12.36\) \\ \midrule

\multirow{2}{*}{50} & \checkmark & \(126.879 \pm 14.657\)  & \(106.552 \pm 27.17\) & \(26.223 \pm 13.973\) \\ 
                    & \xmark & \(121.543 \pm 13.628\)  & \(105.143 \pm 29.365\) & \(27.18 \pm 12.526\) \\ \midrule

\multirow{2}{*}{100} & \checkmark & \(125.656 \pm 13.925\)  & \(105.945 \pm 26.776\) & \(25.594 \pm 12.923\) \\ 
                     & \xmark & \(121.481 \pm 13.483\)  & \(105.466 \pm 29.429\) & \(27.037 \pm 12.245\) \\ \midrule
                               
\multirow{2}{*}{500} & \checkmark & \(126.712 \pm 13.816\) & \(106.884 \pm 28.135\) & \(26.648 \pm 12.113\) \\
                     & \xmark & \(121.491 \pm 13.447\) & \(105.796 \pm 29.63\) & \(27.199 \pm 11.909\) \\ \bottomrule
                   
\end{tabularx}
\end{table}

\begin{table}[H]
\caption{Average percentage of relative pixel changes from the unrefined to the refined segmentation masks on the test dataset. \(S\) represents the number of DDIM sampling steps. The format of the results indicates the mean \(\pm\) std. across 11 testing tiles in one repetition of each experiment.}
\label{tab:stats-percentage-pixel-changes}
\begin{tabularx}{\textwidth}{CCCC}
\toprule
\multirow{2}{*}{\boldmath \(S\) } & \multirow{2}{*}{ \textbf{Noise} } & \multicolumn{1}{c}{\textbf{White}\boldmath \(\to\) \textbf{Black}} & \multicolumn{1}{c}{\textbf{Black} \boldmath\(\to\) \textbf{White}} \\ \cmidrule{3-4}
                         &                         & \textbf{Mean (\%) }& \textbf{Mean (\%)} \\ \midrule
\multirow{2}{*}{10} & \checkmark & \(25.693 \pm 3.287\) & \(0.637 \pm 0.157\) \\ 
                    & \xmark & \(24.43 \pm3.001\) & \(0.631 \pm 0.172\) \\ \midrule

\multirow{2}{*}{50} & \checkmark & \(25.456 \pm 3.31\) & \(0.655 \pm 0.168\) \\ 
                    & \xmark & \( 24.369 \pm 2.979\) & \(0.646 \pm 0.181\) \\ \midrule
                   
\multirow{2}{*}{100} & \checkmark & \(25.196 \pm 2.962\) & \(0.648 \pm 0.182\) \\ 
                     & \xmark & \(24.35 \pm 2.495\) & \(0.826 \pm 0.244\) \\ \midrule
                               
\multirow{2}{*}{500} & \checkmark & \(25.41 \pm 3.033\) & \(0.657 \pm 0.174\) \\
                     & \xmark & \(24.361 \pm 2.967\) & \(0.65 \pm 0.183\) \\ \bottomrule
                   
\end{tabularx}
\end{table}

\begin{adjustwidth}{-\extralength}{0cm}

\reftitle{References}

\PublishersNote{}
\end{adjustwidth}
\end{document}